%% file: arxiv.tex
\documentclass[10pt,twocolumn,letterpaper]{article}

\usepackage{iccv}              


\definecolor{iccvblue}{rgb}{0.21,0.49,0.74}
\usepackage[pagebackref,breaklinks,colorlinks,allcolors=iccvblue]{hyperref}

\usepackage[utf8]{inputenc} 
\usepackage[T1]{fontenc}    
\usepackage{url}            
\usepackage{booktabs}       
\usepackage{amsfonts}       
\usepackage{nicefrac}       
\usepackage{microtype}      
\usepackage{wrapfig}

\usepackage{setspace}

\usepackage{xcolor}         
\usepackage{graphicx} 
\usepackage{tabularx}
\usepackage{amsmath}
\usepackage{xspace}
\usepackage{graphbox}
\usepackage{float}
\usepackage[normalem]{ulem}
\usepackage{multirow}
\usepackage{caption}

\usepackage{makecell} 
\usepackage{algorithm}
\usepackage{algorithmic}

\def\paperID{1728} 
\def\confName{ICCV}
\def\confYear{2025}

\newcommand{\OM}{FreeMorph}
\newcommand{\OMO}{FreeMorph }
\newcommand{\LineNum}[1]{\STATE #1}

\newcommand\yk[1]{{\color{black}#1}}

\title{FreeMorph: Tuning-Free Generalized Image Morphing with Diffusion Model}

\author{Yukang Cao$^{1\ast}$ \quad Chenyang Si$^{2\ast\ddagger}$ \quad Jinghao Wang$^{3}$ \quad Ziwei Liu$^{1\dagger}$\\
$^1$S-Lab, Nanyang Technological University, $^2$Nanjing University\\
$^3$The Chinese University of Hong Kong\\
\url{https://yukangcao.github.io/FreeMorph/}
}

\begin{document}
\twocolumn[{
            \renewcommand\twocolumn[1][]{#1}
            \maketitle
            \vspace{-3em}
            \begin{center}
                \includegraphics[width=\linewidth]{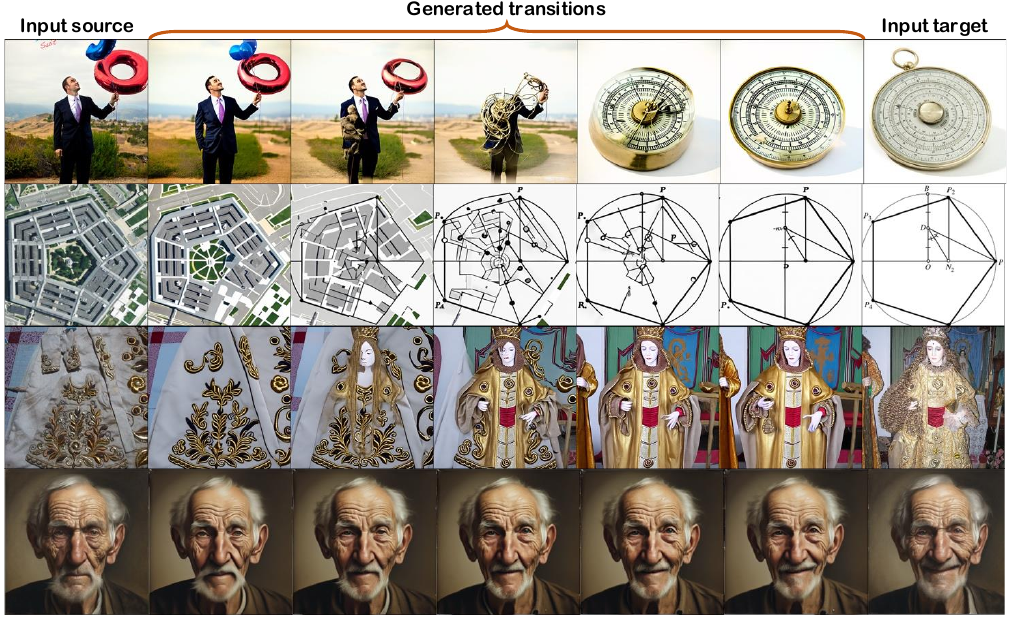}
                \vspace{-2.em}
                \captionof{figure}{\textbf{Examples of image morphing obtained via \OM.} Given two input images, \OMO effectively generates smooth transitions between them within 30 seconds.}
                \label{fig:teaser}
                \vspace{-1em}
            \end{center}
        }]

\renewcommand*{\thefootnote}{\fnsymbol{footnote}}
\footnotetext{$^\ast$ Equal contributions, $^\ddagger$ Project lead, $^\dagger$ Corresponding author.}
\footnotetext{Jinghao was a Master student at NTU during this work.}
\renewcommand*{\thefootnote}{\arabic{footnote}}

\input{Sections/0_abstract}
\input{Sections/1_introduction}

\input{Sections/2_related_works}
\input{Sections/3_methodology}

\input{Sections/4_experiments}

\input{Sections/5_conclusion}

{
    \small
    \bibliographystyle{ieeenat_fullname}
    \bibliography{morphing}
}

\appendix
\onecolumn
\input{Sections/a_supp}

\end{document}

%% file: Sections/0_abstract.tex
\begin{abstract}
    We present \textbf{FreeMorph}, the first tuning-free method for image morphing that accommodates inputs with different semantics or layouts. Unlike existing methods that rely on fine-tuning pre-trained diffusion models and are limited by time constraints and semantic/layout discrepancies, FreeMorph delivers high-fidelity image morphing without requiring per-instance training. 
    Despite their efficiency and potential, tuning-free methods face challenges in maintaining high-quality results due to the non-linear nature of the multi-step denoising process and biases inherited from the pre-trained diffusion model. 
    In this paper, we introduce FreeMorph to address these challenges by integrating two key innovations.
    \textbf{1)} We first propose a \textbf{guidance-aware spherical interpolation} design that incorporates explicit guidance from the input images by modifying the self-attention modules, thereby addressing identity loss and ensuring directional transitions throughout the generated sequence.
    \textbf{2)} We further introduce a \textbf{step-oriented \yk{variation trend}} that blends self-attention modules derived from each input image to achieve controlled and consistent transitions that respect both inputs. 
    Our extensive evaluations demonstrate that FreeMorph outperforms existing methods, being $10\times \sim 50 \times$ faster and establishing a new state-of-the-art for image morphing. 
\end{abstract}

%% file: Sections/1_introduction.tex
\section{Introduction}

Given two distinct input images, image morphing~\citep{zope2017survey, differential_morphing} aims to gradually change attributes such as shape, texture, and overall layout to produce a series of intermediate images that transition smoothly from one to the other.
This process is widely used in fields such as animation, film, and photo editing~\citep{aloraibi2023image, wolberg1996recent, wolberg1998image}, offering an effective means of enhancing creative expression. Historically, image morphing relied on image warping~\citep{smythe1990two, wolberg1990digital, fant1986nonaliasing} for aligning corresponding points and on color interpolation~\citep{beier1992feature, lee1998fast} for blending. These methods, however, often fall short when handling complex textural and semantic transitions, making them less effective for images with intricate details. With advancements in deep learning, Generative Adversarial Networks (GANs)~\citep{goodfellow2014generative, karras2019style, brock2019large, sauer2023stylegan} and Variational Autoencoders (VAEs)~\citep{kingma2013auto} have significantly improved image morphing by enabling latent code interpolation. Despite their capabilities, these approaches still face challenges with real-world images due to limited training data and information loss during GAN inversion. This underscores the need for methods that better preserve identity and offer greater generalization.

Recently, with the availability of large-scale text-image datasets, vision-language models (e.g., Chameleon~\citep{team2024chameleon}), diffusion models (e.g., Stable Diffusion~\citep{stable-diffusion, saharia2022photorealistic, rombach2022high}), and transformers (e.g., PixArt-$\alpha$~\citep{chen2023pixart}, FLUX~\citep{flux}) have demonstrated impressive capabilities in generating high-quality images from text prompts. These advancements have paved the way for new generative image morphing techniques. Specifically, \citet{wang2023interpolating} leverages the local linearity of CLIP-based text embeddings to create smooth transitions by interpolating latent image features. Building on this idea, IMPUS~\citep{yang2023impus} introduces a multi-phase training framework that includes optimizing text embeddings and training Low-Rank Adaptation (LoRA) modules to better capture semantics. While this method yields more visually appealing results, it requires extensive training, typically around 30 minutes per case. DiffMorpher~\citep{zhang2024diffmorpher} proposes to directly interpolate latent noise and leverage Adaptive Instance Normalization (AdaIN) to improve performance. However, these methods still struggle to process images with diverse semantics and intricate layouts, limiting their practical effectiveness.


Given these issues, our objective is to achieve image morphing without requiring further tuning. Nonetheless, this goal introduces two key challenges:
\textbf{1) Non-directional transitions and identity loss\footnote{Non-directional transitions, akin to identity loss, result in generated images that deviate from the identity of the input images.}.} While converting input images into latent features using a pre-trained diffusion model and then applying spherical interpolation might seem straightforward, this approach often results in inconsistent transitions. This is due to the non-linear nature of the multi-step denoising process. Additionally, this method inherits biases from the pre-trained model, which can lead to identity loss in the generated images.
\textbf{2) Achieving consistent transitions\footnote{Inconsistent transitions are those with abrupt changes.}.} A diffusion model does not inherently provide an effective "\yk{variation trend}" to capture the gradual changes between images. Consequently, achieving smooth and gradual transitions in a tuning-free manner remains a significant challenge without additional adjustments.

In this paper, we present \textit{\textbf{\OM}}, a novel tuning-free method capable of instantly generating directional and realistic transitions between two images. Our method introduces two novel components: 
\textbf{1) Guidance-aware spherical interpolation:}
We first enhance the pre-trained diffusion model by incorporating explicit guidance from the input images through modifications to its self-attention modules. 
This is achieved through spherical interpolation, which produces intermediate features used in two key ways. 
First, we perform \textit{spherical feature aggregation} to blend the key and value features of the self-attention modules, ensuring consistent transitions throughout the generated image sequence. 
Second, to address identity loss, we introduce a \textit{prior-driven self-attention mechanism} that incorporates explicit guidance from the input images to preserve their unique identities.
\textbf{2) Step-oriented \yk{variation trend}:} To achieve consistent transitions, we introduce a novel step-oriented \yk{variation trend}. This method blends two self-attention modules, each derived from one of the input images, enabling a controlled and consistent transition that respects both inputs. 
To further improve the quality of the generated image sequences, we designed an improved reverse denoising and forward diffusion process that seamlessly integrates these innovative components into the original DDIM framework. As shown in Fig.~\ref{fig:teaser} and Fig.~\ref{fig:more-result}, our approach adeptly handles diverse input types, whether they have similar or distinct semantics and layouts, producing smooth and realistic transitions.


To thoroughly assess \OMO and benchmark it against current methods, we also collect a new evaluation dataset that includes four distinct sets of image pairs, categorized by their semantic and layout similarity. Our extensive evaluations demonstrate that \OMO substantially outperforms existing approaches. FreeMorph produces high-fidelity image sequences with smooth and coherent transformations in under 30 seconds, making it $\boldsymbol{50 \times}$ faster than IMPUS~\citep{yang2023impus} and $\boldsymbol{10 \times}$ faster than DiffMorpher~\citep{zhang2024diffmorpher}.

%% file: Sections/2_related_works.tex
\section{Related Work}
\label{sec:related}
\paragraph{Text-to-Image Generation.} 
Recently, diffusion models~\citep{rombach2022high,podell2023sdxl,saharia2022photorealistic,ramesh2022hierarchical} have emerged as the \textit{de facto} standard for text-to-image generation. These models employ a series of denoising steps (e.g., DDIM, DDPM)~\citep{ho2020denoising, song2021denoising} to transform Gaussian noise into images, effectively capturing and interpreting details from textual prompts.
Trained on billions of text-image pairs~\citep{schuhmann2022laion}, these models exhibit a remarkable ability to understand the distribution of real-world images, generating high-quality, diverse outputs while maintaining strong generalization capabilities.
Our work harnesses the capabilities of diffusion models, particularly their ability to generate smooth transitions between two specified images~\citep{samuel2024norm, khrulkov2022understanding, he2024aid}, to address the image morphing task.

\vspace{-1em}
\paragraph{Image Morphing.} 
Image morphing is a long-standing computer vision and graphics problem. Before the deep learning era, techniques such as mesh warping~\citep{smythe1990two,wolberg1990digital,fant1986nonaliasing} and field morphing~\citep{beier1992feature,lee1998fast} were the primary approaches in this domain. 
Early approaches~\cite{park2020neural, fish2020image} utilize GANs~\cite{goodfellow2020generative} to achieve this objective. However, they generally suffer from three main limitations: (1) the need for extensive training, (2) poor generalization to out-of-domain inputs, and (3) an inability to handle inputs with varying layouts and semantic structures.
Recently, advancements in diffusion models have led to significant progress, as demonstrated by methods such as DiffMorpher~\citep{zhang2024diffmorpher}, IMPUS~\citep{yang2023impus}, and the work of~\citet{wang2023interpolating}. These approaches focus on optimizing text embeddings for two images and fine-tuning pre-trained text-to-image diffusion models to achieve smooth interpolation.
However, they often require extensive fine-tuning for each image pair and are limited to images with similar semantics and layouts. This can also hinder the generalizability of pre-trained diffusion models due to constraints imposed by LoRA modules in the U-Net architecture.
In contrast, our method offers a tuning-free framework that requires no modifications to the original diffusion models, thereby preserving their inherent generalizability. Additionally, our approach significantly improves efficiency and can handle images with different layouts and semantics, addressing a key limitation of existing techniques.

\vspace{-1em}
\paragraph{Tuning-Free Text-Guided Image Editing.} 
Recent image translation methods have emerged that edit either generated or real-world images using text in a training-free manner, without altering the internal computations of the U-Net. For instance, SDEdit~\citep{meng2021sdedit} proposes a straightforward method that adds $T$ time steps of Gaussian noise to an original image and then denoises it using guiding text.
Conversely, EDICT~\citep{wallace2023edict} and FPI~\citep{meiri2023fixed} focus on inverting a reference image back to the latent space and subsequently applying the inverted latent as a condition guided by text. 
Additionally, methods like P2P~\citep{hertz2022prompt}, PnP~\citep{tumanyan2023plug}, and MasaCtrl~\citep{cao2023masactrl} modify the attention mechanism within diffusion models to enhance alignment between the guiding text and the consistency of generated images with their originals.
Drawing inspiration from these techniques, our method facilitates image morphing in a tuning-free manner. Notably, our approach also achieves comparable image editing performance by framing text-guided editing as a special case of morphing between a real and a generated image.

%% file: Sections/3_methodology.tex
\section{Methodology}
\label{sec:methodology}
Given two independent images, $\mathcal{I}_{\text{left}}$ and $\mathcal{I}_{\text{right}}$, as input, our objective is to generate a sequence of intermediate images $\mathcal{S} = \{\mathcal{I}_j\}_{j=1}^{J}$ that smoothly transforms from one to the other in a tuning-free manner. We set $J = 5$ for the experiments reported in this paper. As illustrated in Algorithm~\ref{alg:algorithm1}, our pipeline employs a pre-trained diffusion model as its foundation and integrates guidance from the input images into the multi-step denoising process. In the subsequent sections, we first introduce the preliminaries that underpin our method in Sec.~\ref{subsec: preliminary}. 
Next, we describe the \OMO framework in detail. 
This framework comprises three main components: 1) the guidance-aware spherical interpolation (Sec.~\ref{subsec:guidance-aware}), which includes our proposed spherical feature aggregation and prior-driven self-attention mechanism; 2) a step-oriented \yk{variation trend} that enables controlled and consistent image morphing (Sec.~\ref{subsec:motion-flow}); and 3) our improved forward diffusion and reverse denoising processes (Sec.~\ref{subsec:steps}).

\subsection{Preliminaries}
\label{subsec: preliminary}

\paragraph{Denoising Diffusion Implicit Model (DDIM).}
\yk{The Denoising Diffusion Implicit Model (DDIM)~\citep{song2021denoising}, trained on large-scale text-image datasets, is designed to reconstruct images from noisy inputs. After training, it establishes a deterministic mapping from an initial noise state $x_T$ to an image $x_0$, a process we refer as reverse denoising steps:}
\begin{equation}
\label{eq:get_image}
\small
\begin{split}
    x_{t-1} = &\sqrt{\bar\alpha _{t-1}}(\frac{x_t-\sqrt{1-\bar\alpha_{t}}\epsilon_\theta(x_t)}{\sqrt{\bar\alpha_{t}}}) \\ + &\sqrt{1-\bar\alpha_{t-1}-\sigma _t^2}\epsilon_{\theta}(x_t, t) + \sigma _t\epsilon _t.
\end{split}
\end{equation}
\yk{Conversely, by inverting the formula above, we can derive the forward diffusion process, which incrementally adds noise to an image to predict its noise state:}
\begin{equation}
\label{eq:get_noise}
\small
\begin{split}
    x_t = &\sqrt{\frac{\bar\alpha _t}{\bar\alpha _{t-1}}}x_{t-1} + \\ &\sqrt{\alpha _t} (\sqrt{\frac{1}{\alpha_t}-1 } - \sqrt{\frac{1}{\alpha_{t-1} } -1})\epsilon _{\theta}(x_{t-1}, t-1).
\end{split}
\end{equation}
\vspace{-2em}



\input{Figures/replace-kv}

\paragraph{Latent Diffusion Model (LDM).} 
Building upon DDIM, the Latent Diffusion Model (LDM)~\citep{rombach2022high} is a refined variant of diffusion models that effectively balances image quality with denoising efficiency. 
Specifically, LDM utilizes a pre-trained variational auto-encoder (VAE)~\citep{kingma2013auto} to map images into a latent space and then trains the diffusion model within this space. 
Furthermore, LDM enhances the UNet architecture by incorporating self-attention modules, cross-attention layers, and residual blocks to integrate text prompts as conditional inputs during image generation. The attention mechanism in LDM's UNet can be formulated as:
\begin{equation}
\mathtt{ATT}(Q, K, V) = \text{softmax}(\frac{Q \cdot K^T}{\sqrt{d_k}}) \cdot V
\label{eq:att}
\end{equation}
where $Q$ denotes the query features from spatial data, and $K$ and $V$ are key and value features derived from either spatial data (for self-attention) or text embeddings (for cross-attention). The noise estimator in LDM is then extended to $\epsilon_\theta(\mathbf{x}_t, t, y)$, where $y$ denotes the text embedding.

Our approach builds upon the Stable Diffusion model~\citep{stable-diffusion}, a pre-trained LDM developed by StabilityAI, and utilizes a vision-language model (VLM), LLaVA~\citep{liu2024llavanext}, for generating captions for the input images.

\subsection{Guidance-aware spherical interpolation}
\label{subsec:guidance-aware}
Existing image morphing methods~\citep{differential_morphing, zhang2024diffmorpher, yang2023impus} typically involve training Low-rank Adaptation (LoRA) modules for each input image to enhance semantic comprehension and achieve smooth transitions. However, this approach is often inefficient and time-consuming and struggles with images that differ in semantics or layout.
In this paper, we propose a tuning-free image morphing approach built on the pre-trained Stable Diffusion model. 
By leveraging the capabilities of DDIM (as in Eq.~\ref{eq:get_noise}) for image inversion and interpolation, one might consider converting the input images ($\mathcal{I}_{\text{left}}$, $\mathcal{I}_{\text{right}}$) into latent features ($\mathbf{z}_{0-\text{left}}$, $\mathbf{z}_{0-\text{right}}$) and applying spherical interpolation may seem like a simple straightforward solution:
\begin{equation}
    \mathbf{z}_{0-j} = \frac{\text{sin}((1 - j) \cdot \phi)}{\text{sin}\phi} \cdot \mathbf{z}_{0-\text{left}} + \frac{\text{sin}(j \cdot \phi)}{\text{sin}\phi} \cdot \mathbf{z}_{0-\text{right}},
\end{equation}
where $j \in [1, J]$ is the index of intermediate images, and $\phi = \text{arccos}(\frac{\mathbf{z}^{T}_{0-\text{left}} \cdot \mathbf{z}_{0-\text{right}}}{||\mathbf{z}_{0-\text{left}}|| \cdot ||\mathbf{z}_{0-\text{right}}})$. Recall that we set $J=5$ in our paper.
However, directly inverting these interpolated latent features $\mathbf{z}_{0-j}$ to generate images often results in inconsistent transitions and identity loss (see Fig.~\ref{fig:replace-kv}). This issue arises because (1) the multi-step denoising process is highly non-linear, leading to discontinuous image sequences, and (2) there is no explicit guidance to control the denoising, causing the model to inherit biases from the pre-trained diffusion model.


\input{Figures/noise-distortion}
\vspace{-1em}
\paragraph{Spherical Feature Aggregation.}
Drawing insights from previous image editing techniques~\citep{cao2023masactrl, hertz2022prompt, parmar2023zero, shi2023dragdiffusion, tumanyan2023plug}, we observed that using the features $\mathbf{z}_{0-j}$ as initialization and replacing the key and value features ($K$ and $V$) in the attention mechanism (as described in Eq.~\ref{eq:att}) with features from the right image $\mathcal{I}_{\text{right}}$ can largely enhance the smoothness and identity preservation of the image transitions, although some imperfections may remain (see Fig.~\ref{fig:replace-kv}). Motivated by this finding, and recognizing that the query features ($Q$) largely reflect the overall image layout, we propose first blending features from both the left and right images ($\mathcal{I}_{\text{left}}$, $\mathcal{I}_{\text{right}}$) to provide explicit guidance for the multi-step denoising process. Specifically, in the denoising step $t$, we first feed the latent of the input images $\mathbf{z}_{t-\text{left}}$ and $\mathbf{z}_{t-\text{right}}$ to the pre-trained UNet $\epsilon_\theta$ to obtain the key and value features. Following that, We then substitute the original $K$ and $V$ with those derived from the input images and compute their average to modify the attention mechanism:
\begin{equation}
\small
\begin{split}
    \mathtt{ATT}(Q_{t-j}, K_{t-j}, V_{t-j}) :&= \frac{1}{2} \cdot (\mathtt{ATT}(Q_{t-j}, K_{t-\text{left}}, V_{t-\text{left}}) \\ &+ \mathtt{ATT}(Q_{t-j}, K_{t-\text{right}}, V_{t-\text{right}})) 
    \label{eq:forward-gradual-forward}
\end{split}
\end{equation}
where $Q_{t-j}$, $K_{t-j}$, $V_{t-j}$ are obtained by inputting $\mathbf{z}_{t-j}$ to the pre-trained UNet $\epsilon_\theta$. Note that $\mathbf{z}_{t-j}$, $\mathbf{z}_{t-\text{left}}$ and $\mathbf{z}_{t-\text{right}}$ are derived based on Eq.~\ref{eq:att}.

\paragraph{Prior-driven Self-attention Mechanism.}
While our feature blending technique significantly improves identity preservation in image morphing, we found that using this approach uniformly in both forward diffusion and reverse denoising stages can result in transitions where the image sequences change minimally and fail to accurately represent the input images (see Fig.~\ref{fig:ablation-gradual}). 
This outcome is anticipated because the latent noise will largely influence the reverse denoising process, as shown in Fig.~\ref{fig:noise-distortion}. Consequently, applying our feature blending, depicted in Eq.~\ref{eq:forward-gradual-forward}, introduces ambiguity as the consistent and strong constraints from the input images cause each latent noise $i$ to appear similar, thereby limiting the effectiveness of the transitions.
To tackle this issue, we further propose a prior-driven self-attention mechanism that prioritizes the latent features from spherical interpolation to ensure smooth transitions within the latent noise, while emphasizing the input images to maintain identity preservation afterward. Specifically, during the reverse denoising stage, we use the approach described in Eq.~\ref{eq:forward-gradual-forward}, while for the forward diffusion steps, we employ a different attention mechanism as follows by modifying the self-attention modules:
\begin{equation}
    \mathtt{ATT}(Q_{t-j}, K_{t-j}, V_{t-j}) := \frac{1}{J}\sum_{k=1}^{J} \mathtt{ATT}(Q_{t-j}, K_{t-k}, V_{t-k})
    \label{eq:forward-gradual-inverse}
\end{equation}
Refer to Sec.~\ref{subsec:ablation} for detailed ablation studies on this design.
\input{Tables/algorithm}

\vspace{-1em}
\subsection{Step-oriented \yk{variation trend}}
\label{subsec:motion-flow}
After obtaining image sequences that are directional and accurately reflect the input identities, the next challenge is to achieve a consistent and gradual transition from the left image $\mathcal{I}_{\text{left}}$ to the right image $\mathcal{I}_{\text{right}}$.
This problem stems from the lack of a "\yk{variation trend}" that captures the changes from $\mathcal{I}_{\text{left}}$ to $ \mathcal{I}_{\text{right}}$. 
To this end, we propose a step-oriented \yk{variation trend} that gradually changes the influence between the input images ($\mathcal{I}_{\text{left}}$ and $\mathcal{I}_{\text{right}}$):
\begin{equation}
\small
\begin{split}
    \mathtt{ATT}(Q_{t-j}, K_{t-j}, V_{t-j}) :&= (1 - \alpha_j) \cdot \mathtt{ATT}(Q_{t-j}, K_{t-left}, V_{t-left}) \\&+ \alpha_j \cdot \mathtt{ATT}(Q_{t-j}, K_{t-right}, V_{t-right}),
    \label{eq:aggressive}
\end{split}
\end{equation}
where $\alpha_j = j / (J + 2 - 1)$, with $J + 2$ representing the total number of images, which includes the $J$ generated images and the 2 input images.

\input{Tables/quantitative}
\input{Figures/more-result}
\subsection{Forward diffusion and reverse denoising process}
\label{subsec:steps}

\paragraph{High-frequency Gaussian Noise Injection.}
As discussed earlier, FreeMorph incorporates features from both the left and right images during the forward diffusion and reverse denoising stages. 
Nevertheless, we have observed that this can occasionally impose overly stringent constraints on the generation process. To mitigate this issue and allow for greater flexibility, we propose introducing Gaussian noise into the latent vector $\mathbf{z}$ in the high-frequency domain after the forward diffusion steps:
\begin{equation}
    \mathbf{z} :=
    \begin{cases} 
        \mathtt{IFFT}(\mathtt{FFT}(\mathbf{z})),  & \mbox{if }\mathbf{m}\mbox{ = 1} \\
        \mathtt{IFFT}(\mathtt{FFT}(\mathbf{\mathbf{g}})), & \mbox{if }\mathbf{m}\mbox{ = 0}
    \end{cases}
\end{equation}
Here, $\mathtt{IFFT}(\cdot)$ and $\mathtt{FFT}(\cdot)$ denote the inverse fast Fourier transform and fast Fourier transform, respectively. $\mathbf{g} \sim \mathcal{N}(0,1)$ represents a randomly sampled noise vector, and $\mathbf{m}$ is a binary high-pass filter mask of the same size as $\mathbf{z}$.

\vspace{-1em}
\paragraph{Overall process.}
To enhance the efficacy of our image morphing process, we have found that consistently applying either guidance-aware spherical interpolation (Sec.~\ref{subsec:guidance-aware}) or step-oriented \yk{variation trend} (Sec.~\ref{subsec:motion-flow}) across all denoising steps yields suboptimal results (see Sec.~\ref{subsec:ablation}). To address this, we have developed a refined approach for both forward diffusion and reverse denoising processes. We provide an overview algorithm of our proposed FreeMorph in Algorithm.~\ref{alg:algorithm1}. Specifically:
\begin{itemize}
    \item \textbf{Forward diffusion:} We use the standard self-attention mechanism for the first $\lambda_1 \cdot T$ steps. From $\lambda_1 \cdot T$ to $\lambda_2 \cdot T$, we apply the feature blending technique from Eq.~\ref{eq:forward-gradual-inverse}. For the remaining steps, we implement the step-oriented \yk{variation trend}.
    \item \textbf{Reverse denoising:} We begin with the step-oriented \yk{variation trend} for the first $\lambda_3 \cdot T$ steps, followed by the feature blending method from Eq.~\ref{eq:forward-gradual-forward} for steps between $\lambda_3 \cdot T$ and $\lambda_4 \cdot T$. The process ends with the original self-attention mechanism for the final steps to produce images with higher fidelity.
\end{itemize}
Here, $\lambda_1$, $\lambda_2$, $\lambda_3$, and $\lambda_4$ are hyper-parameters and $T=50$ is the total number of steps.

%% file: Figures/replace-kv.tex
\begin{figure*}[t]
  \centering
   \includegraphics[width=\linewidth]{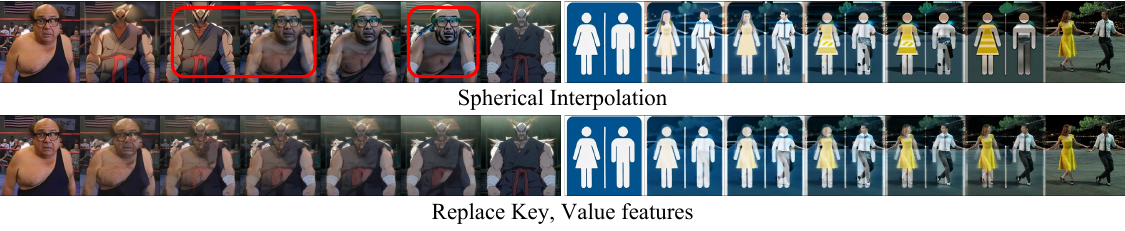}
   \vspace{-2.em}
  \caption{\textbf{Replacing the key and value feature in the attention mechanism.} We can observe that good key and value features would lead to smooth transitions and identity preservation.}
   \label{fig:replace-kv}
   \vspace{-2em}
\end{figure*} 

%% file: Figures/noise-distortion.tex
\begin{figure*}[t]
  \centering
   \includegraphics[width=\linewidth]{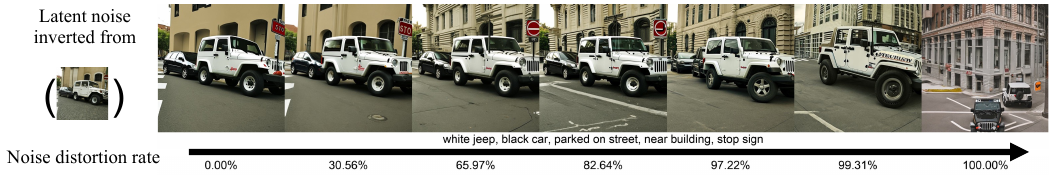}
   \vspace{-2.em}
  \caption{\textbf{Effectiveness of the latent noise on the generated images.} The pre-trained diffusion model is robust to the noise distortion within the latent space.}
   \label{fig:noise-distortion}
   \vspace{-2em}
\end{figure*} 

%% file: Tables/algorithm.tex
\begin{algorithm}[t]
    \caption{FreeMorph}
    \textbf{Input:} $\mathcal{I}_{\text{left}}$, $\mathcal{I}_{\text{right}}$
    \begin{algorithmic}
    
    \LineNum{1:} Caption the input images via pre-trained LLaVA  $\rightarrow \text{Text}_{\text{left}}$, $\text{Text}_{\text{right}}$.
        
    \LineNum{2:} Obtain image features $\mathbf{z}_{0-\text{left}}$, $\mathbf{z}_{0-\text{right}}$, and text embedding $y_{\text{left}}$, $y_{\text{right}}$ via VAE and text encoder of pre-trained Stable Diffusion.
    
    \LineNum{3:} Applying spherical interpolation to obtain $\mathbf{z}_{0-j}$ where $j \in [1, J]$ as initialization.
    
    \LineNum{4:} Forward diffusion steps (from image to latent noise):
    \end{algorithmic}

    \begin{algorithmic}
        \FOR {$t=1$ to $T$}
        \IF {$t < \lambda_1 \cdot T$ }
        \STATE{Apply the original attention mechanism.}
        \ELSIF{$t < \lambda_2 \cdot T$}
        \STATE{Apply the prior-driven self-attention mechanism as in Eq.~\ref{eq:forward-gradual-inverse}.}
        \ELSE
        \STATE{Apply the step-oriented motion flow as in Eq.~\ref{eq:aggressive}.}
        \ENDIF
        \ENDFOR
    \end{algorithmic}
    
    \begin{algorithmic}
        \LineNum{5:} High-frequency Gaussian noise injection.
        \LineNum{6:} Reverse denoising steps (from latent noise to image):
    \end{algorithmic}
    
    \begin{algorithmic}
        \FOR {$t=1$ to $T$}
        \IF {$t < \lambda_3 \cdot T$ }
        \STATE{Apply the step-oriented motion flow as in Eq.~\ref{eq:aggressive}.}
        \ELSIF{$t < \lambda_4 \cdot T$}
        \STATE{Apply the spherical feature aggregation as in Eq.~\ref{eq:forward-gradual-forward}.}
        \ELSE
        \STATE{Apply the original attention mechanism.}
        \ENDIF
        \ENDFOR
    \end{algorithmic}
    
    \begin{algorithmic}
    \LineNum{7:} Add text-conditioned features.
    \end{algorithmic}
    
    \textbf{Output:} $J$ intermediate images gradually change from $\mathcal{I}_{\text{left}}$ to $\mathcal{I}_{\text{right}}$.

\label{alg:algorithm1}
\end{algorithm}

%% file: Tables/quantitative.tex
\begin{table*}[t]
\centering
\caption{\textbf{Quantitative comparison with existing image morphing techniques.}}
\vspace{-1em}
\label{tab:quantitative}
\resizebox{1.\textwidth}{!}{
\begin{tabular}{l||ccc||ccc||ccc}

\toprule[1pt]
{\multirow{2}*{Method}} & \multicolumn{3}{c||}{MorphBench} & \multicolumn{3}{c||}{Morph4Data} & \multicolumn{3}{c}{Overall} \\

{} & {$\text{LPIPS}_\text{sum} \downarrow$} & {$\text{FID}_\text{mean}\downarrow$} &  {$\text{PPL}_\text{sum}\downarrow$} & {$\text{LPIPS}_\text{sum} \downarrow$} & {$\text{FID}_\text{mean}\downarrow$} & {$\text{PPL}_\text{sum}\downarrow$} & {$\text{LPIPS}_\text{sum} \downarrow$} & {$\text{FID}_\text{mean}\downarrow$} & {$\text{PPL}_\text{sum}\downarrow$}\\
\midrule[1pt]
IMPUS~\cite{yang2023impus}   & 130.52 & 152.43 & 3263.03 & 134.88 & 210.66 & 3199.90  & 265.40 & 174.76 & 6462.93 \\         
DiffMorpher~\cite{zhang2024diffmorpher}  & 90.57 & 157.18 & 2264.20 & 98.56 & 292.54 & 2394.05 & 189.13 & 209.10 & 4658.25 \\
Spherical Interpolation   & 119.77 & 169.17 & 2994.35 & 103.74 & 245.22  & 2593.58 & 223.52 & 198.34 & 5587.93 \\
Ours  & \textbf{84.91} & \textbf{141.32}  & \textbf{2122.80} & \textbf{80.30} & \textbf{201.09} & \textbf{2007.52} & \textbf{162.99} & \textbf{152.88} & \textbf{4192.82}\\
\midrule[1pt]
\end{tabular}
}
\vspace{-1em}

\end{table*}

%% file: Figures/more-result.tex
\begin{figure*}[t]
  \centering
   \includegraphics[width=\linewidth]{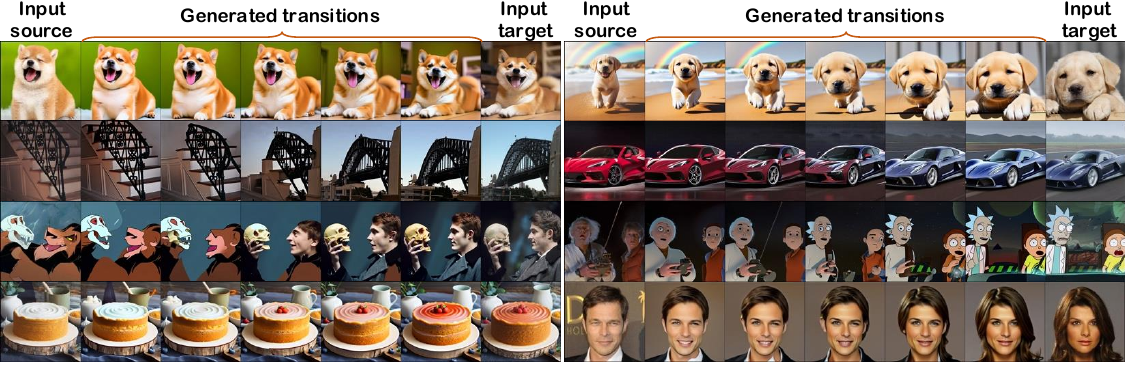}
   \vspace{-2.em}
  \caption{\textbf{More results produced by \OM.} Our method can achieve smooth and high-fidelity image transitions for input images with either similar or different semantics and layouts.}
   \label{fig:more-result}
   \vspace{-2em}
\end{figure*} 

%% file: Sections/4_experiments.tex
\vspace{-0.5em}
\section{Experiments}
\label{sec:exp}

We evaluate the performance of FreeMorph across various scenarios, comparing it with state-of-the-art image morphing techniques and conducting ablation studies to highlight the effectiveness of our proposed components.

\vspace{-1em}
\paragraph{Implementation Details.}
We use version 2.1 of the publicly available Stable Diffusion model. Both the forward diffusion and reverse denoising processes employ a DDIM schedule with $T = 50$ steps. It takes under 30 seconds for our method to produce a morphing sequence using NVIDIA A100 GPU. Following the Stable Diffusion setup, we operate on an image resolution of $768 \times 768$. We set the classifier-free guidance (CFG) parameter to 7.5 to incorporate text-conditioned features. The hyperparameters are set as follows: $\lambda_1 = 0.3$, $\lambda_2 = 0.6$, $\lambda_3 = 0.2$, $\lambda_4 = 0.6$.

\vspace{-1em}
\paragraph{Evaluation Datasets.}
DiffMorpher~\citep{zhang2024diffmorpher} introduced MorphBench, which includes 24 animation pairs and 66 image pairs, predominantly featuring images with similar semantics or layouts. To complement this dataset and mitigate potential biases, we introduce \textbf{Morph4Data}, a newly curated evaluation dataset comprising four categories: 
1) \textit{Class-A}, consisting of 25 image pairs with similar layouts but different semantics, sourced from \citet{wang2023interpolating}; 
2) \textit{Class-B}, containing image pairs with both similar layouts and semantics, including 11 pairs of faces from CelebA-HQ~\citep{karras2017progressive} and 10 pairs of various car types; 
3) \textit{Class-C}, featuring 15 pairs of randomly sampled images from ImageNet-1K~\citep{deng2009imagenet} with no semantic or layout similarity; 
4) \textit{Class-D}, comprising 15 pairs of dog and cat images randomly sampled from the internet.

\input{Figures/comparison}
\subsection{Quantitative Evaluations}
\label{subsec:quantitative}
Following IMPUS~\citep{yang2023impus} and DiffMorpher~\citep{zhang2024diffmorpher}, 
we conducted quantitative comparisons using the following metrics: 1) Frechet Inception Distance (FID)~\citep{heusel2017gans}, which assesses the similarity between the distributions of input and generated images; 2) Perceptual Path Length (PPL)~\citep{karras2020analyzing}, where we calculate the sum of PPL loss between adjacent images; and 3) Learned Perceptual Image Patch Similarity (LPIPS)~\citep{zhang2018perceptual}, which we also sum for adjacent images to evaluate the smoothness and coherence of the generated transitions. The results, detailed in Table~\ref{tab:quantitative}, demonstrate the superior performance of our method across both datasets, showing enhanced fidelity, smoothness, and directness.

\paragraph{User studies}
\yk{To enhance our comparative analysis by including human preferences, we conducted user studies. We recruited 30 volunteers, including animators, AI experts, and gaming enthusiasts aged 20 to 35, to select their preferred results. Each participant was shown 50 random pairs of comparative results. The outcomes, presented in Table~\ref{tab:user-studies}, demonstrate the subjective effectiveness of our proposed approach. Note that \textit{slerp} denotes the method that only applies spherical interpolation.}
\input{Tables/user-studies}

\subsection{Qualitative Evaluations}
\label{subsec:qualitative}

\input{Tables/quantitative_ablation}
\paragraph{Qualitative Results.} In Fig.~\ref{fig:teaser} and Fig.~\ref{fig:more-result}, we present a wide range of results produced by \OM, which consistently demonstrate its ability to generate high-quality and smooth transitions. \OMO excels across diverse scenarios, accommodating images with different semantics and layouts, as well as those with similar characteristics. FreeMorph also effectively handles subtle variations, such as cakes with different colors and individuals with different expressions.

\vspace{-1em}
\paragraph{Qualitative Comparisons.} We provide qualitative comparisons with existing image morphing methods in Fig.~\ref{fig:comparison}. \yk{An effective image morphing outcome should exhibit gradual transitions from the source (left) image to the target (right) image while preserving the original identities. Based on this criterion, several observations can be made:} \textbf{1)} When handling images with varying semantics and layouts, IMPUS~\citep{yang2023impus} exhibits identity loss and produces unsmooth transitions; \yk{For instance, in the second example of Fig.~\ref{fig:comparison}, IMPUS exhibits (i) identity loss, where the third generated image deviates from the original identity, and (ii) an abrupt transition between the third and fourth generated images.} \textbf{2)} Although Diffmorpher~\citep{zhang2024diffmorpher} achieves smoother transitions than IMPUS, its results often suffer from blurriness and lower overall quality (see the first example in Fig.~\ref{fig:comparison}); \textbf{3)} We also evaluate a baseline approach, \yk{`\textit{Slerp}'}, which involves applying only spherical interpolation and the DDIM process. The visualizations show that this baseline approach struggles with (i) accurately interpreting the input images due to the absence of explicit guidance, (ii) suboptimal image quality, \yk{and (iii) abrupt transitions}. In contrast, our method consistently delivers superior performance, characterized by smoother transitions and higher image quality. Additional comparisons are available in the Appendix.

\vspace{-1em}
\subsection{Further Analysis}
\label{subsec:ablation}

\input{Figures/ablation-gradual}
\paragraph{Analysis of Guidance-aware Spherical Interpolation.}
In Fig.~\ref{fig:ablation-gradual}, we present ablation studies to evaluate the effects of the proposed spherical feature aggregation (Eq.~\ref{eq:forward-gradual-forward}) and the prior-driven self-attention mechanism (Eq.~\ref{eq:forward-gradual-inverse}). The results indicate that using either component alone produces suboptimal outcomes. Specifically, (i) spherical feature aggregation is crucial for achieving directional transitions in which the characteristics of $\mathcal{I}_{\text{left}}$ gradually diminish, and (ii) the prior-driven self-attention mechanism is vital for preserving identity in the generated images. The combination of both components allows FreeMorph to produce smooth transitions while effectively maintaining identity. By comparing the last two rows in Fig.~\ref{fig:ablation-gradual}, we demonstrate the importance of our step-oriented \yk{variation trend} and the specially designed reverse and forward processes.

\input{Figures/ablation-design}
\vspace{-1.5em}
\paragraph{Analysis of Reverse and Forward Process.}
In Fig.~\ref{fig:ablation-design}, we evaluate our method against two variants: (i) “Ours (Var-A),” which omits the original attention mechanism, and (ii) “Ours (Var-B),” which swaps the application steps of the guidance-aware spherical interpolation and the step-oriented \yk{variation trend} in both the reverse and forward processes. A comparison of these variants with our final design reveals that (i) the original attention mechanism is crucial for achieving high-fidelity results, and (ii) the specific configuration of the reverse and forward processes in our final design yields optimal performance.

\vspace{-1.5em}
\paragraph{Analysis of Step-oriented \yk{Variation Trend}.}
In Fig.~\ref{fig:ablation2}, we first disable the proposed step-oriented \yk{variation trend} to assess its impact. We observe that without this component, the model tends to produce abrupt changes rather than smooth transitions. Additionally, the final generated image exhibits high-contrast colors that differ from the target image $\mathcal{I}_{\text{right}}$. In contrast, the step-oriented \yk{variation trend} enables our method to achieve smoother transitions and produce a final image that is more closely aligned with the target image.

\input{Figures/ablation2}
\vspace{-1.5em}
\paragraph{Analysis of High-frequency Noise Injection.}
We then disable high-frequency noise injection and present the corresponding ablation study in Fig.~\ref{fig:ablation2}. The results indicate that incorporating the proposed high-frequency noise injection enhances the model’s flexibility and contributes to smoother transitions.

%% file: Figures/comparison.tex
\begin{figure*}[t]
  \centering
   \includegraphics[width=\linewidth]{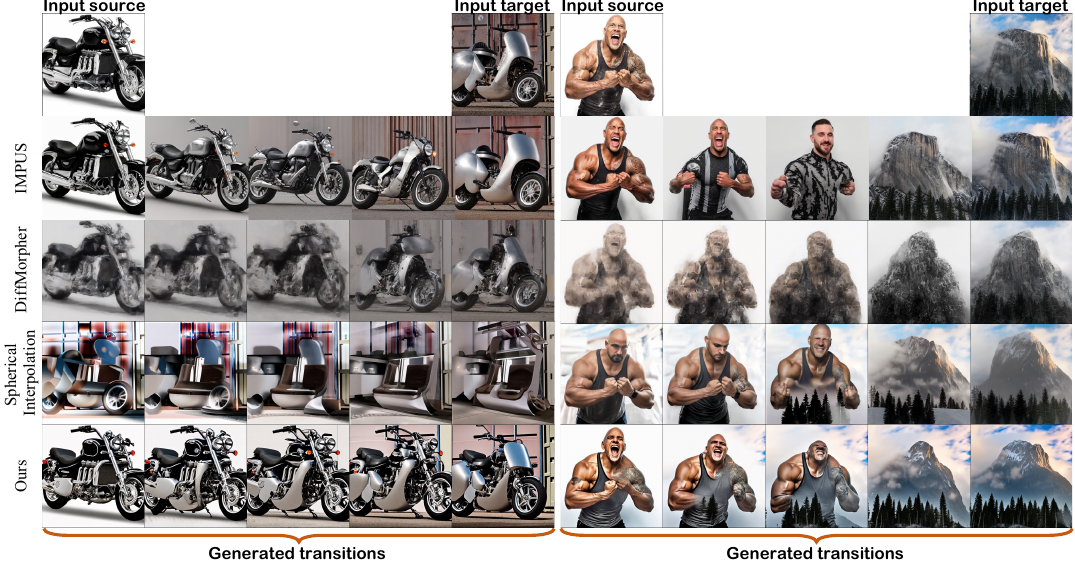}
   \vspace{-2.em}
  \caption{\textbf{Qualitative comparison with existing image morphing techniques.} Unlike other methods that struggle or fail to generate smooth and high-fidelity results without identity loss, our approach consistently achieves high-quality transitions, yielding superior results.}
   \label{fig:comparison}
   \vspace{-1em}
\end{figure*} 

%% file: Tables/user-studies.tex
\begin{table}[t]
\centering
\vspace{-1em}
\caption{\textbf{User studies.}}
\vspace{-1em}
\label{tab:user-studies}
\resizebox{0.47\textwidth}{!}{
\begin{tabular}{l||cccc}

\toprule[1pt]
{} & IMPUS~\cite{yang2023impus} & DiffMorpher~\cite{zhang2024diffmorpher} & \textit{Slerp} & Ours \\

{Preference} & 17.16$\%$ & $14.89\%$ & $7.82\%$ & $\textbf{60.13\%}$ \\
\midrule[1pt]
\end{tabular}
}
\vspace{-2em}

\end{table}

%% file: Tables/quantitative_ablation.tex
\begin{table*}[t]
\centering
\caption{\textbf{Quantitative comparison for ablation studies.}}
\vspace{-1em}
\label{tab:quantitative-ablation}
\resizebox{1.\textwidth}{!}{
\begin{tabular}{l||ccc||ccc||ccc}

\toprule[1pt]
{\multirow{2}*{Method}} & \multicolumn{3}{c||}{MorphBench} & \multicolumn{3}{c||}{Morph4Data} & \multicolumn{3}{c}{Overall} \\

{} & {$\text{LPIPS}_\text{sum} \downarrow$} & {$\text{FID}_\text{mean}\downarrow$} &  {$\text{PPL}_\text{sum}\downarrow$} & {$\text{LPIPS}_\text{sum} \downarrow$} & {$\text{FID}_\text{mean}\downarrow$} & {$\text{PPL}_\text{sum}\downarrow$} & {$\text{LPIPS}_\text{sum} \downarrow$} & {$\text{FID}_\text{mean}\downarrow$} & {$\text{PPL}_\text{sum}\downarrow$}\\
\midrule[1pt]
w/ only Eq.~\ref{eq:forward-gradual-inverse}   & 157.01 & 320.05 & 3425.19 & 141.12 & 411.80 & 3028.05 & 298.13 & 355.24 & 6453.24 \\         
w/ only Eq.~\ref{eq:forward-gradual-forward}   & 99.69 & 155.51 & 2491.10 & 90.80 & 217.26 & 2270.05 & 190.49 & 179.20 & 4761.15\\
w/ only Eq.~\ref{eq:forward-gradual-inverse} and Eq.~\ref{eq:forward-gradual-forward}  & 211.52 & 243.08 & 5288.10 & 139.55 & 290.11 & 3488.87 & 351.08 & 261.12 & 8776.96 \\
w/o noise injection  & 99.49 & 154.53 & 2487.16 & 89.12 & 211.23 & 2228.03 & 188.61 & 176.28 & 4715.19 \\
w/o Eq.~\ref{eq:forward-gradual-forward}  & 87.41 & 155.46 & 2185.30 & 81.10 & 218.95 & 2027.58 & 168.52 & 179.82 & 4212.88 \\
w/o Eq.~\ref{eq:forward-gradual-inverse}   & 120.01 & 148.54 & 3000.35 & 101.28 & 215.43 & 2572.06 & 221.30 & 174.19 & 5572.41\\
w/o step-oriented motion flow    & 118.50 & 154.71 & 2962.48 & 93.39 & 214.93 & 2334.68 & 211.89 & 177.80 & 5297.17\\
Ours (Var-A)   & 153.40 & 184.54 & 3835.08 & 115.91 & 243.20 & 2897.63 & 269.31 & 207.04 & 6732.70\\
Ours (Var-B)   & 93.54 & 158.44 & 2338.62 & 85.76  & 245.36 & 2144.08 & 179.31 & 191.78 & 4482.70\\
\midrule[1pt]
Ours  & \textbf{84.91} & \textbf{141.32}  & \textbf{2122.80} & \textbf{80.30} & \textbf{201.09} & \textbf{2007.52}  & \textbf{162.99} & \textbf{152.88} & \textbf{4192.82}\\
\midrule[1pt]
\end{tabular}
}
\vspace{-1em}
\end{table*}

%% file: Figures/ablation-gradual.tex
\begin{figure}[t]
  \centering
  \vspace{-1em}
   \includegraphics[width=\linewidth]{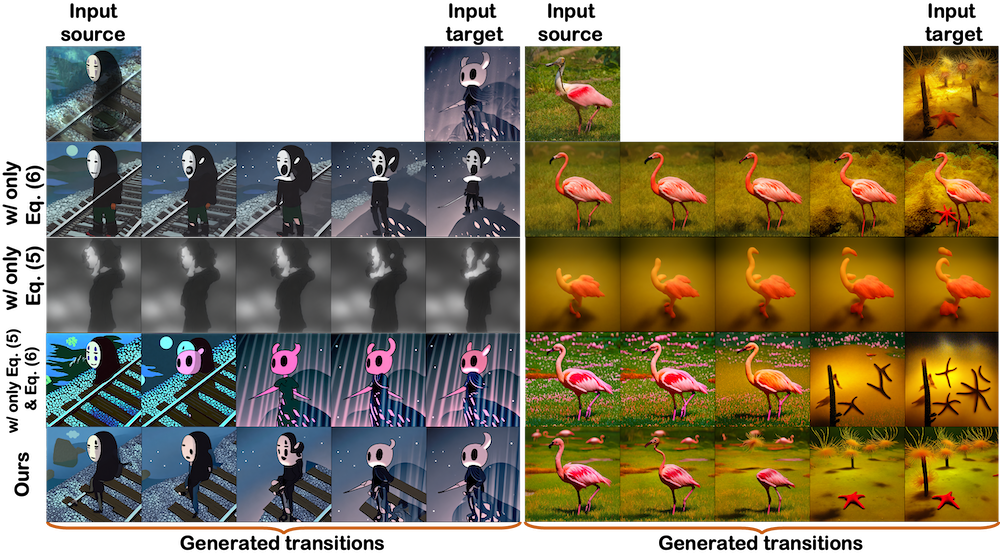}
   \vspace{-2.em}
  \caption{\textbf{Analysis of guidance-aware spherical interpolation.}}
   \label{fig:ablation-gradual}
   \vspace{-1em}
\end{figure} 

%% file: Figures/ablation-design.tex
\begin{figure}[t]
  \centering
   \includegraphics[width=\linewidth]{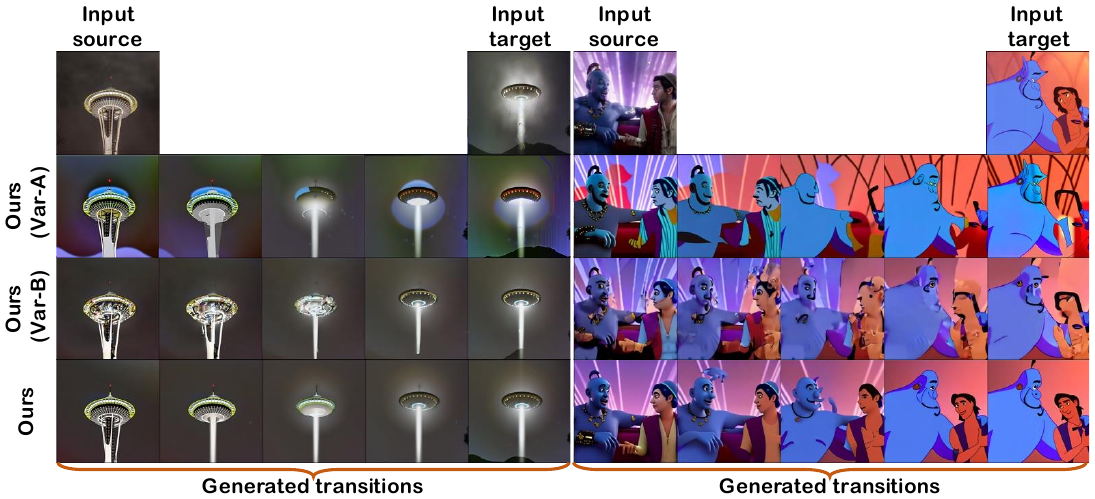}
   \vspace{-2.em}
  \caption{\textbf{Analysis of reverse diffusion and forward denoising process.}}
   \label{fig:ablation-design}
   \vspace{-1.5em}
\end{figure} 

%% file: Figures/ablation2.tex
\begin{figure}[t]
  \centering
  \vspace{-1em}
   \includegraphics[width=\linewidth]{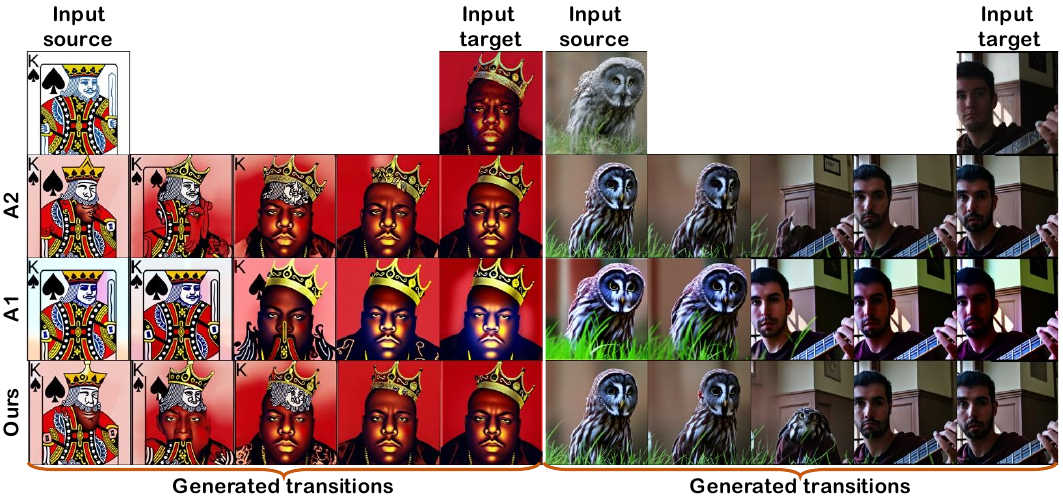}
   \vspace{-2em}
  \caption{\textbf{Analysis of high-frequency noise injection and step-oriented motion flow.} \textbf{A1:} w/o step-oriented motion flow; \textbf{A2:} w/o high-frequency noise injection}
   \label{fig:ablation2}
   \vspace{-2em}
\end{figure} 

%% file: Sections/5_conclusion.tex
\vspace{-0.5em}
\section{Conclusion}
\vspace{-0.5em}
We have introduced \OM, a novel tuning-free pipeline capable of generating smooth, high-quality transitions between two input images in under 30 seconds. Specifically, we propose incorporating explicit guidance from the input images by modifying the self-attention modules. This is achieved through two novel components: spherical feature aggregation and a prior-driven self-attention mechanism. Additionally, we introduce a step-oriented \yk{variation trend} to ensure directional transitions consistent with both input images. We also designed an improved forward diffusion and reverse denoising process to integrate our proposed modules into the original DDIM framework. Extensive experiments demonstrate that FreeMorph delivers high-fidelity results across various scenarios, significantly outperforming existing image morphing techniques.


%% file: Sections/a_supp.tex

\section{Further Analysis}

\subsection{Usage of the Fast Fourier Transform (FFT)}

\yk{In our approach, we employ the fast Fourier transform (FFT) to inject high-frequency Gaussian noise, which enhances flexibility. An alternative and straightforward variation involves replacing the FFT with the discrete cosine transform (DCT). To investigate this, we conducted experiments using both FFT and DCT, presenting the results in Fig.~\ref{fig:dct}. The findings indicate that DCT performs comparably to FFT.
}
\input{Figures/supp/analysis-dct}

\newpage

\section{Qualitative Comparisons}
\subsection{Qualitative Comparisons with AID~\cite{he2024aid} and Smooth Diffusion~\cite{guo2024smooth}}
\yk{
In addition to the comparisons discussed in the main paper, we extend our evaluation to include AID~\cite{he2024aid} and Smooth Diffusion~\cite{guo2024smooth}. As illustrated in Fig.~\ref{fig:aid} and Fig.~\ref{fig:smooth-diffusion}, the results demonstrate that both methods are limited to processing images with similar layouts and semantics, rendering them ineffective for inputs with different layouts or semantics. Beyond their qualitative shortcomings, it is worth noting that \textbf{(1)} AID relies on IP-Adapter for image morphing, which adversely affects training efficiency, and \textbf{(2)} Smooth Diffusion requires parameter tuning, making it slower and less efficient than our approach.
}
\input{Figures/supp/qualitative-aid}
\input{Figures/supp/qualitative-smooth-diffusion}

\newpage

\subsection{Comparison with video generative models}
Given the rapid development of video generative techniques. Methods like PixelDance~\cite{zeng2024make} and SEINE~\cite{chen2023seine} have been designed to achieve image morphing. We hereby provide more comparisons with these video generative models to demonstrate our performance. Considering PixelDance hasn't released code or an online demo, we ran FreeMorph on the examples from their webpages to perform qualitative comparisons (see Fig.~\ref{fig:video} below). Surprisingly, our method performs similarly with PixelDance and outperforms SEINE in reducing ghost artifacts.

\begin{figure}[!h]
\centering
\includegraphics[width=\linewidth]{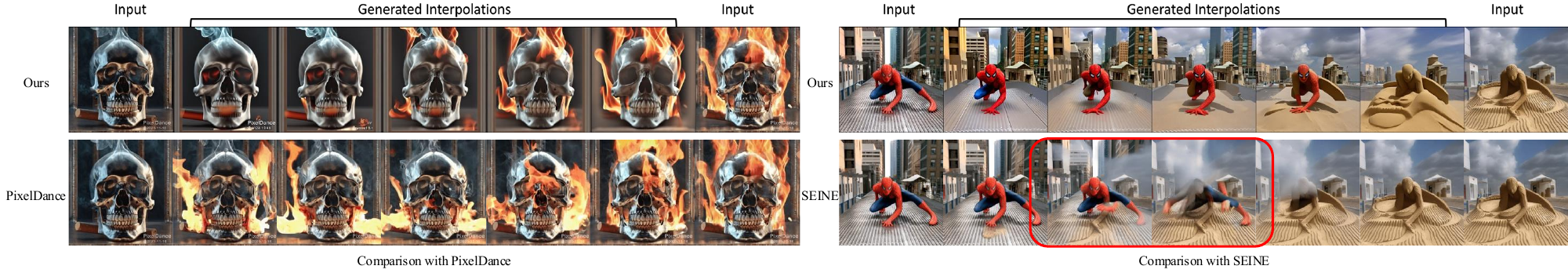}
\caption[Comparisons with video generative models.]{\textbf{Comparisons with video generative models.}}
\label{fig:video}
\end{figure}

\subsection{Comparison with GAN-based morphing methods}
\begin{figure}[!h]
\centering
\includegraphics[width=\linewidth]{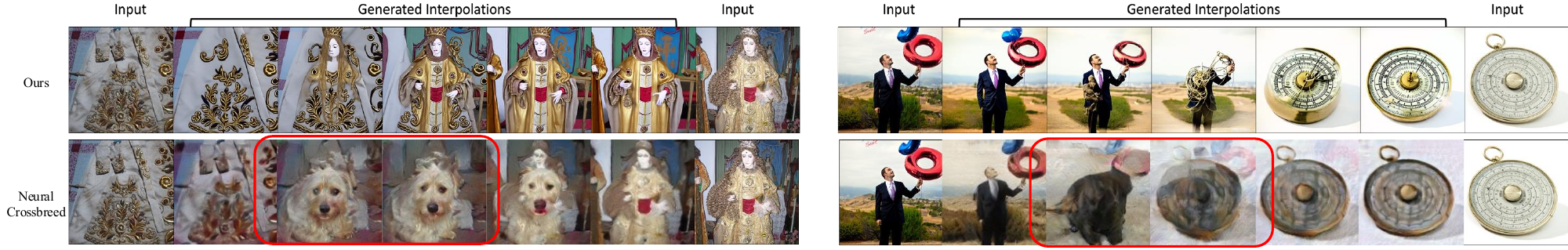}
\caption[Comparisons with GAN-based morphing methods.]{\textbf{Comparison with GAN-based morphing methods.}}
\label{fig:gan}
\end{figure}

We further compare our method with the early GAN-based morphing method (Neural Crossbreed) to demonstrate the performance. The results, presented in Fig.~\ref{fig:gan}, show superior image quality, identity preservation, and smoother transitions. Unlike GAN-based approaches, ours is training-free, is able to handle out-of-domain inputs, and remains robust to varying layouts and semantics. Additional evaluations and discussions will be included in the revised version.

\newpage
\subsection{Comparison with Wang and Golland~\cite{wang2023interpolating}}
\begin{figure}[!h]
\centering
\includegraphics[width=\linewidth]{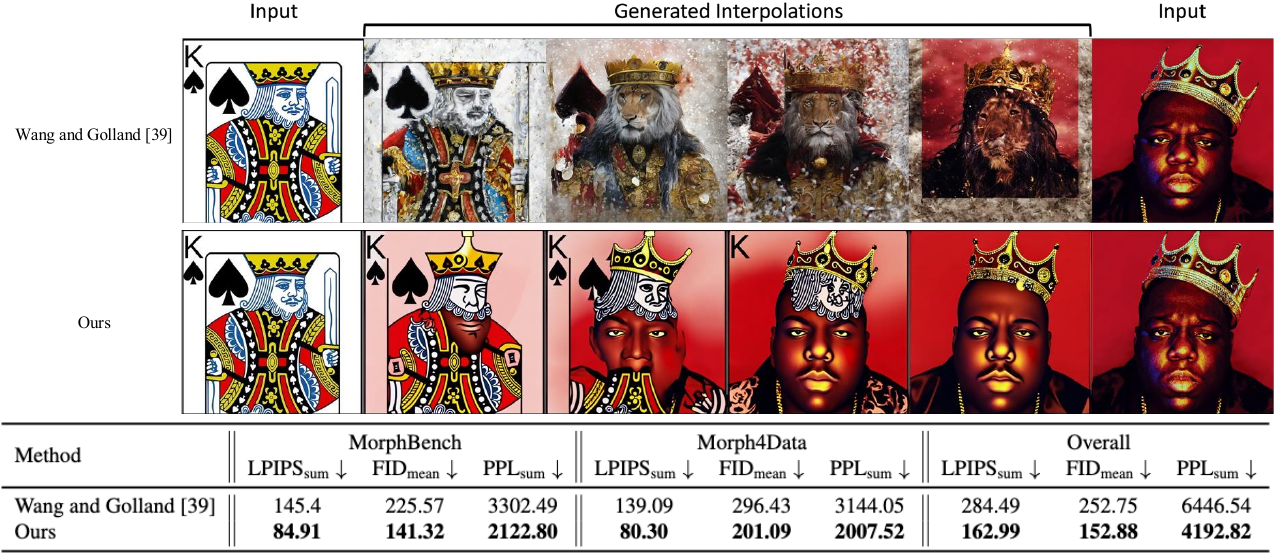}
\caption[Comparison with Wang and Golland~\cite{wang2023interpolating}]{\textbf{Comparison with Wang and Golland~\cite{wang2023interpolating}.}}
\label{fig:wang-and-golland}
\end{figure}

We further compare with Wang and Golland [39] and present the results in Fig.~\ref{fig:wang-and-golland}. We can clearly observe that our method consistently show superior performance over it, both qualitatively and quantitatively.

\subsection{Experiments with different poses/actions}
\begin{figure}[!h]
\centering
\includegraphics[width=\linewidth]{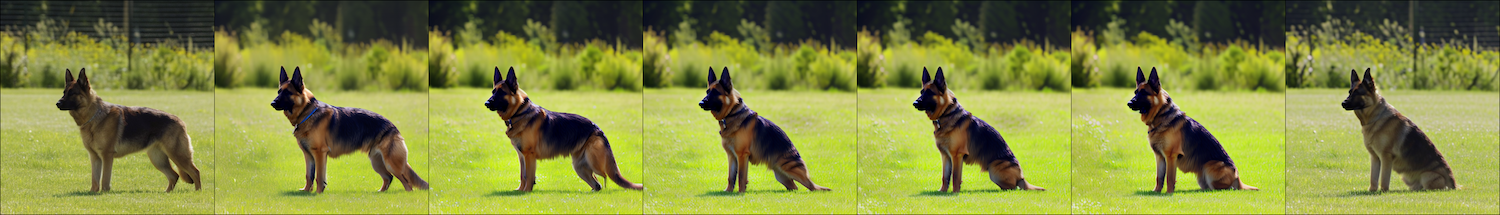}
\caption[Qualitative results with different poses/actions]{\textbf{Qualitative results with different poses/actions.}}
\label{fig:poses}
\end{figure}

We further present results for various poses and actions below (Fig.~\ref{fig:poses}), using input images from the MorphBench dataset.

\newpage

\subsection{Additional Qualitative Comparisons}
We provide additional qualitative comparisons with three methods in Fig.~\ref{fig:more-comparison1}–Fig.~\ref{fig:more-comparison8}. These results reinforce the conclusions drawn in Sec. 4.2 of the main paper, offering further evidence of the superior performance of our \OMO method in achieving high-fidelity and smooth image morphing.
\input{Figures/supp/qualitative-comparison}

\clearpage

\section{More Qualitative Results}
To provide a better understanding of the intermediate generated transitions, in addition to the animated videos, we also present generated images in Fig.~\ref{fig:more-results1}–Fig.~\ref{fig:more-results4}, \textbf{which correspond to the animated videos in the HTML file}.

\input{Figures/supp/qualitative-results}

\clearpage
\section{Visualization of Morph4Data}
We present a range of visualizations from our collected Morph4Data to enhance understanding of the dataset and the distinctions among its different classes.
\input{Figures/supp/morph4data}

\section{Applications}
We highlight that our \OMO method can be adapted for image editing tasks. Specifically, this is accomplished by (1) using the same image as both the "input source" and "input target," and (2) employing different text prompts, where the first prompt describes the original image and subsequent prompts indicate the desired editing direction.
An example is provided in Fig.~\ref{fig:editing}. 
Notably, our method produces image editing results that align correctly with the text prompts, preserving the original identity while effectively generating smooth transitions throughout the editing process.

\input{Figures/supp/application}

\clearpage
\section{Limitations and Failure Cases}
While our method establishes a new state-of-the-art, we acknowledge that it has certain limitations. We illustrate several failure cases in Fig.~\ref{fig:failure}. Specifically: \textbf{1)} Although our model can achieve reasonable results when processing images with no semantic or layout similarity, the generated transitions may not be smooth, potentially leading to abrupt changes. \textbf{2)} Our method inherits biases from Stable Diffusion~\citep{stable-diffusion}, resulting in difficulties in accurately transitioning images that model human limbs.

\input{Figures/failure}

\section{Societal Impact}

Our research advances the image morphing task across a range of semantics and layouts, establishing a more versatile pipeline. However, there is a risk of misuse, such as brands creating misleading advertisements that distort consumer perceptions and create unrealistic product expectations. This practice not only undermines consumer trust but also raises significant ethical concerns about the authenticity of marketing. Additionally, the complexities of copyright and consent are amplified, as manipulated images blur the lines of ownership and accountability. Therefore, we advocate for strict legal compliance and usage restrictions to regulate the application of image morphing techniques and derivative models.

%% file: Figures/supp/analysis-dct.tex
\begin{figure}[h]
  \centering
   \includegraphics[width=\linewidth]{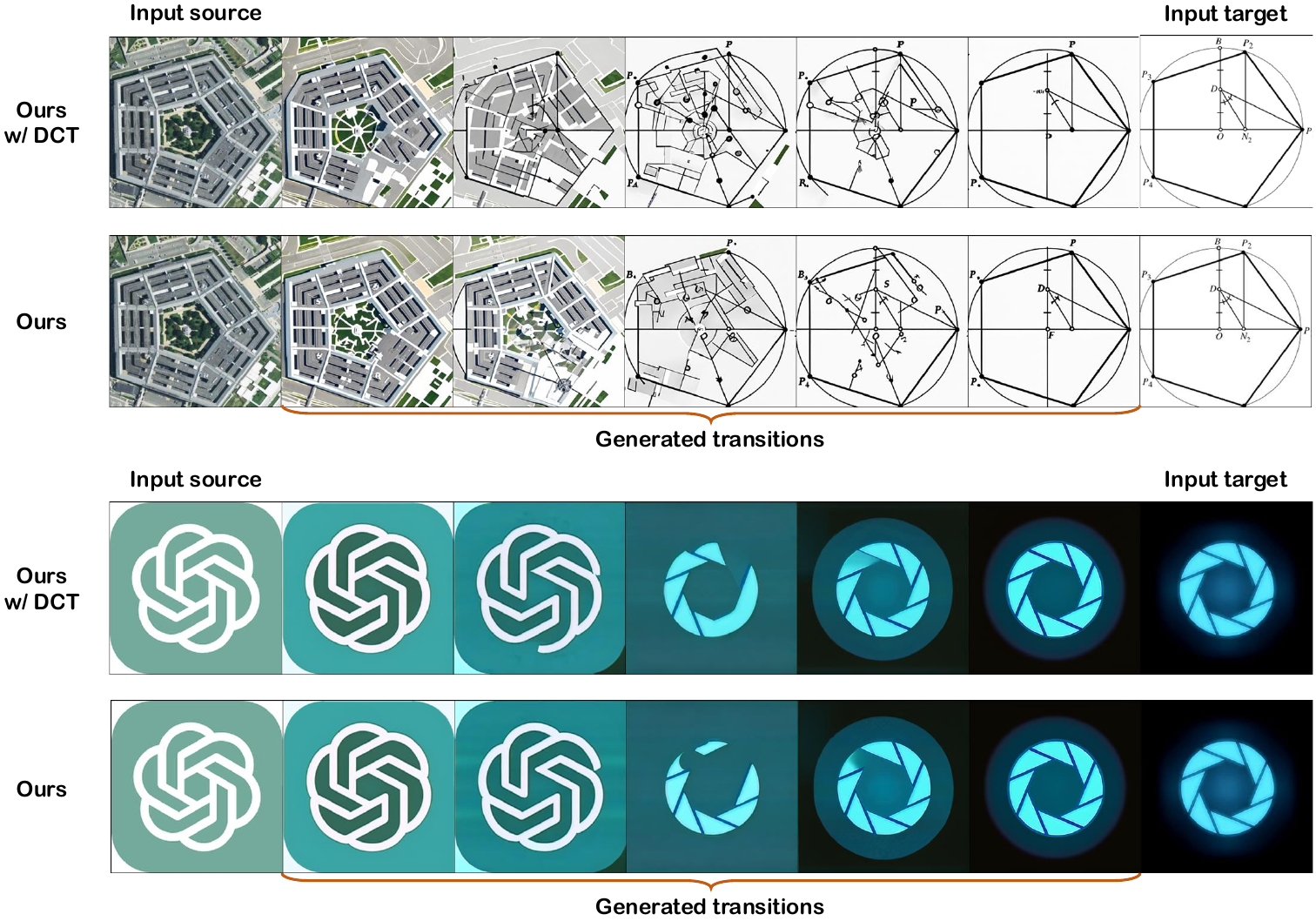}
  \caption[Analysis of the usage of Fast Fourier Transform (FFT) over Discrete Cosine Transform (DCT).]{\textbf{Analysis of the usage of Fast Fourier Transform (FFT) over Discrete Cosine Transform (DCT).}}
   \label{fig:dct}
\end{figure} 

%% file: Figures/supp/qualitative-aid.tex
\begin{figure}[h]
  \centering
   \includegraphics[width=\linewidth]{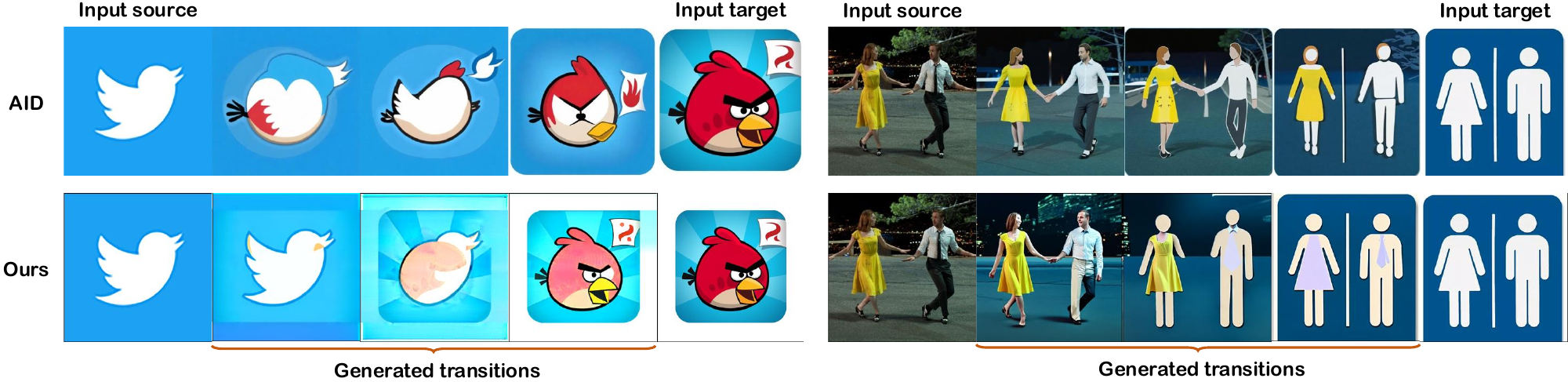}
  \caption[Qualitative comparisons with AID~\cite{he2024aid}.]{\textbf{Qualitative comparisons with AID~\cite{he2024aid}.}}
   \label{fig:aid}
\end{figure} 

%% file: Figures/supp/qualitative-smooth-diffusion.tex
\begin{figure}[h]
  \centering
  \vspace{-2em}
   \includegraphics[width=0.9\linewidth]{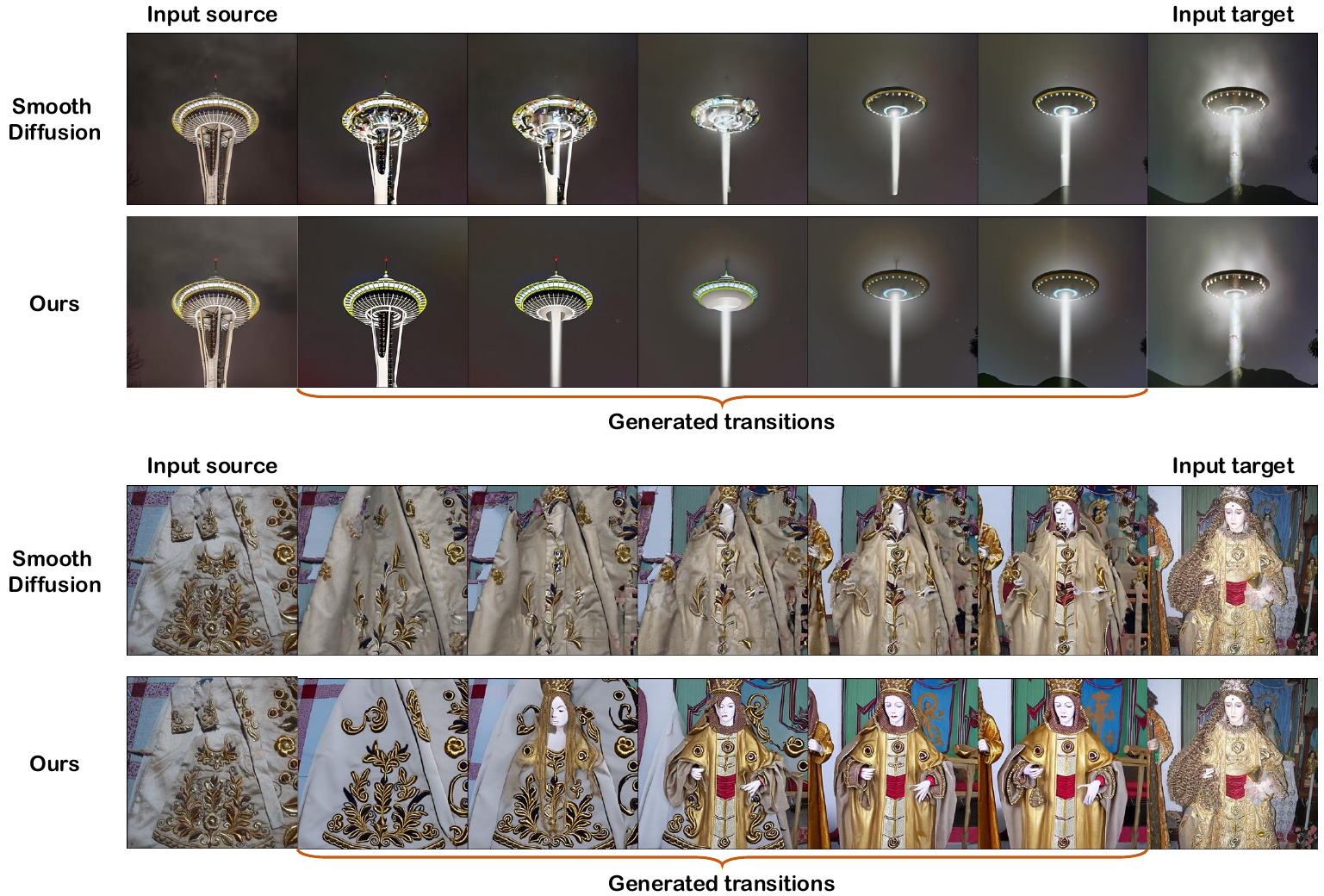}
  \caption[Qualitative comparisons with Smooth Diffusion~\cite{guo2024smooth}]{\textbf{Qualitative comparisons with Smooth Diffusion~\cite{guo2024smooth}}}
   \label{fig:smooth-diffusion}
   \vspace{-2em}
\end{figure} 

%% file: Figures/supp/qualitative-comparison.tex
\begin{figure*}[h]
  \centering
  \vspace{-1em}
   \includegraphics[width=0.8\linewidth]{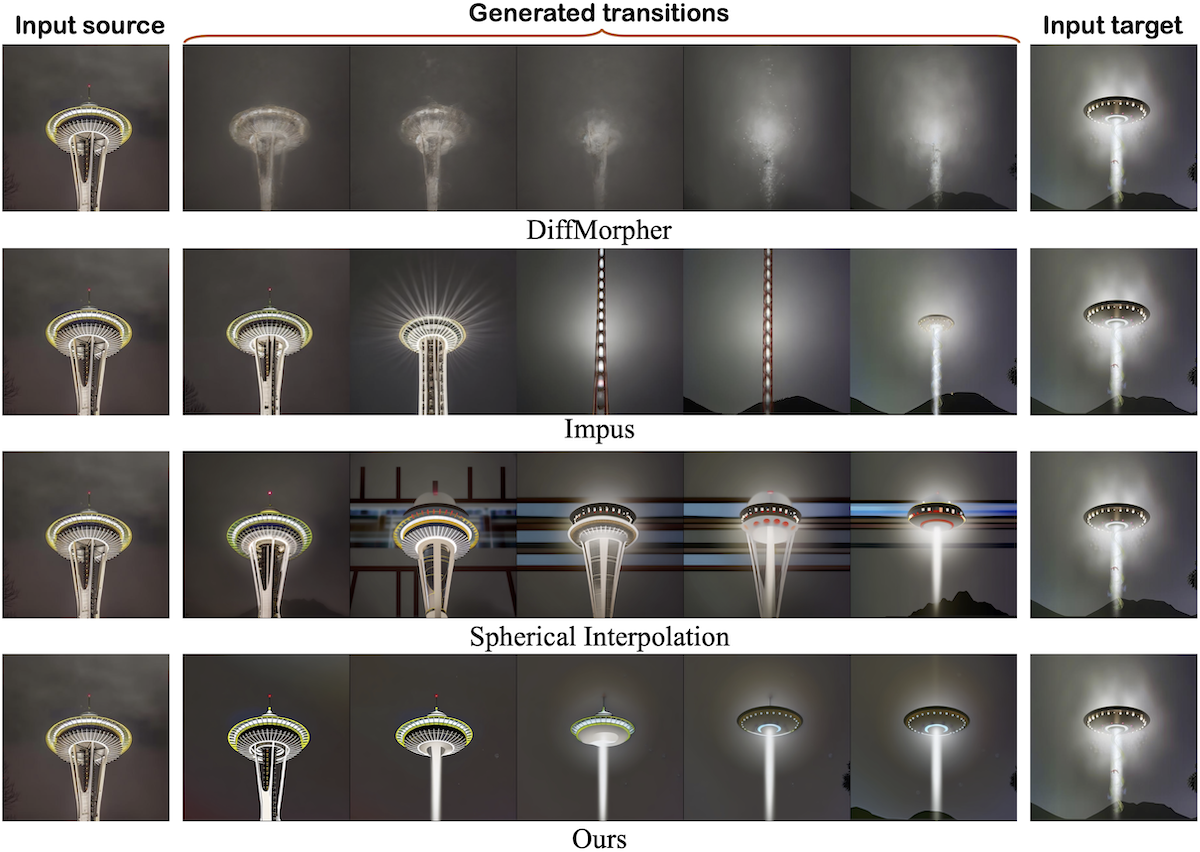}
   \includegraphics[width=0.8\linewidth]{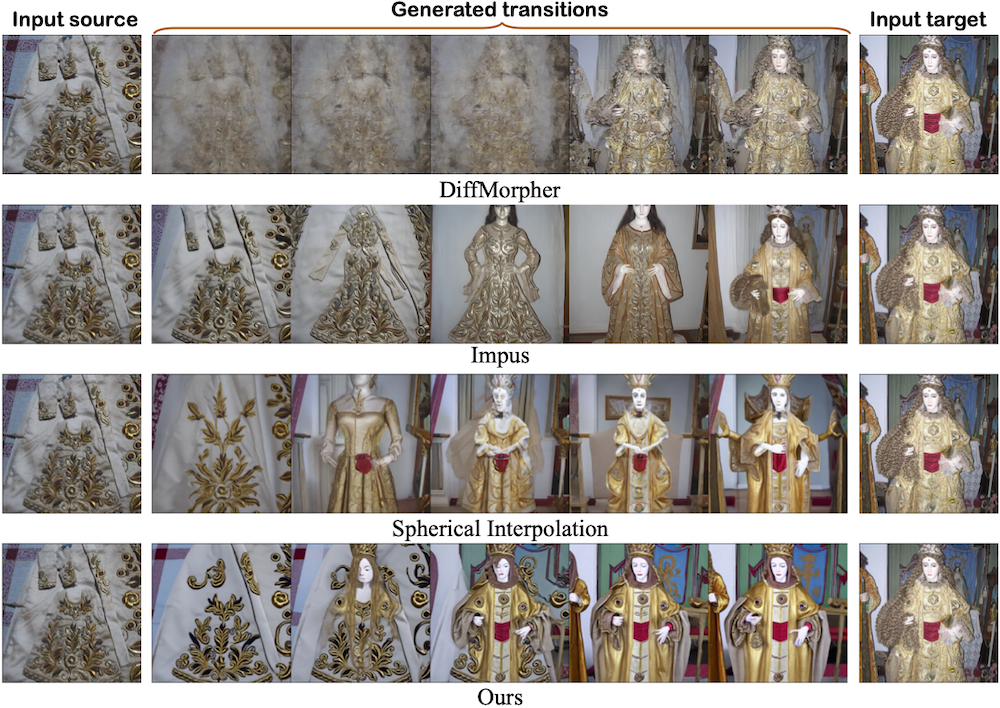}
  \caption[More qualitative comparisons with existing techniques (Part I).]{\textbf{More qualitative comparisons with existing techniques (Part I).} }
   \label{fig:more-comparison1}
\end{figure*} 

\begin{figure*}[h]
  \centering
   \includegraphics[width=0.9\linewidth]{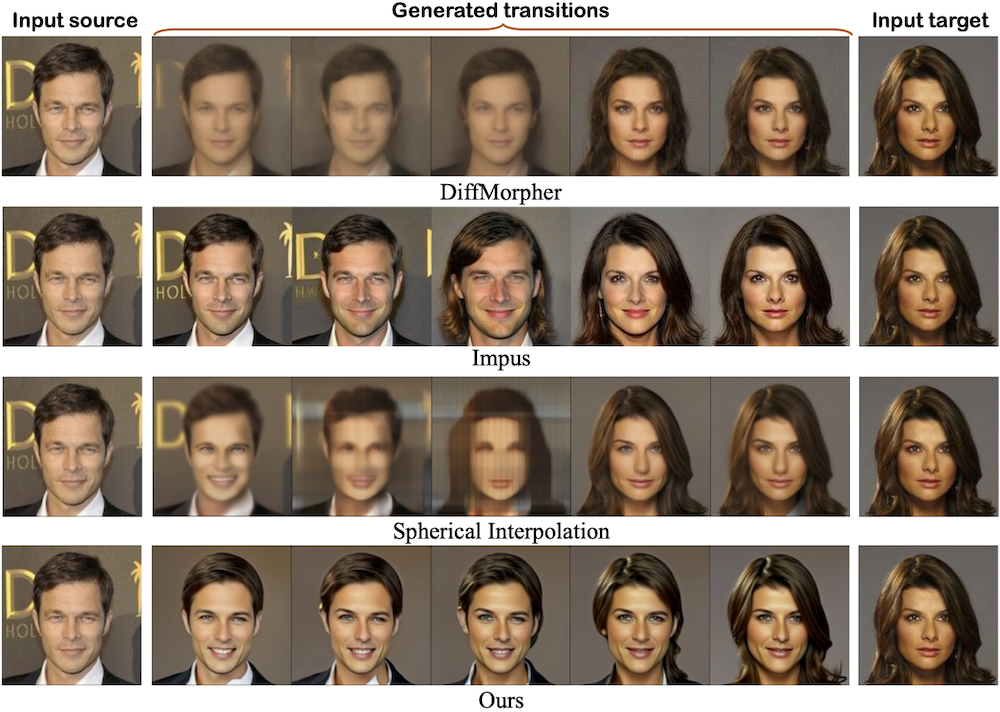}
   \includegraphics[width=0.9\linewidth]{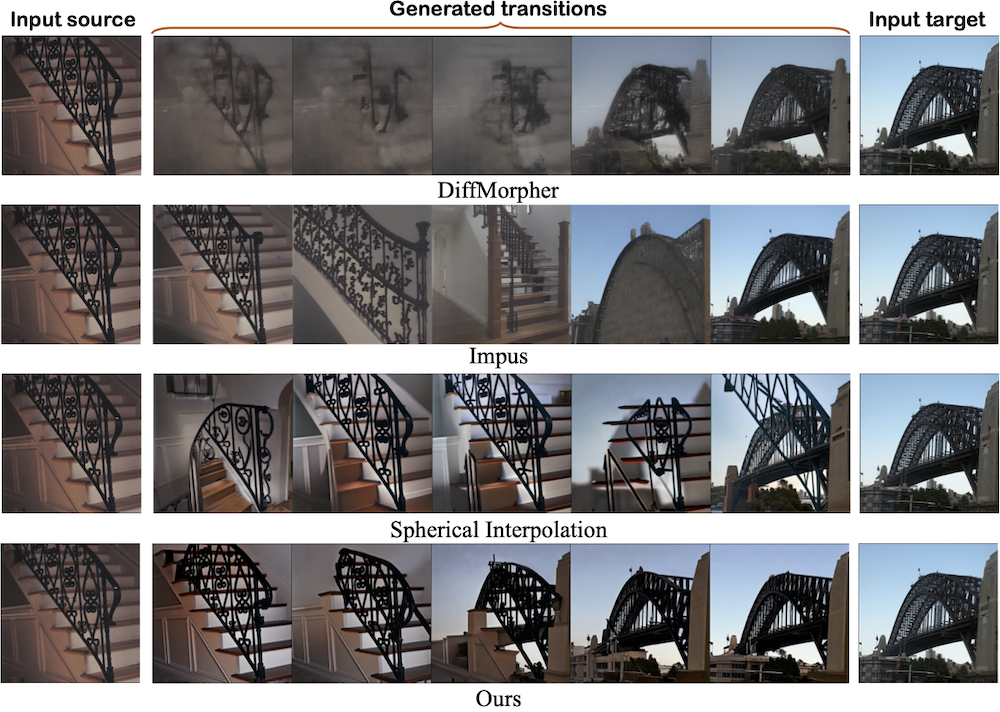}
  \caption[More qualitative comparisons with existing techniques (Part II).]{\textbf{More qualitative comparisons with existing techniques (Part II).} }
   \label{fig:more-comparison2}
\end{figure*} 

\begin{figure*}[h]
  \centering
   \includegraphics[width=0.9\linewidth]{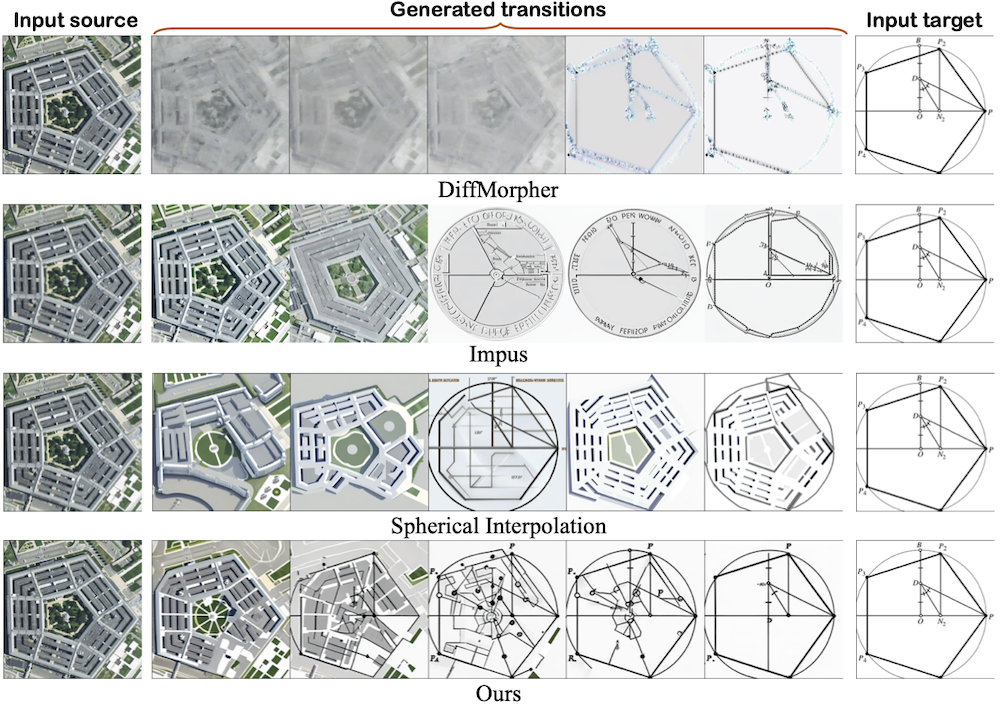}
   \includegraphics[width=0.9\linewidth]{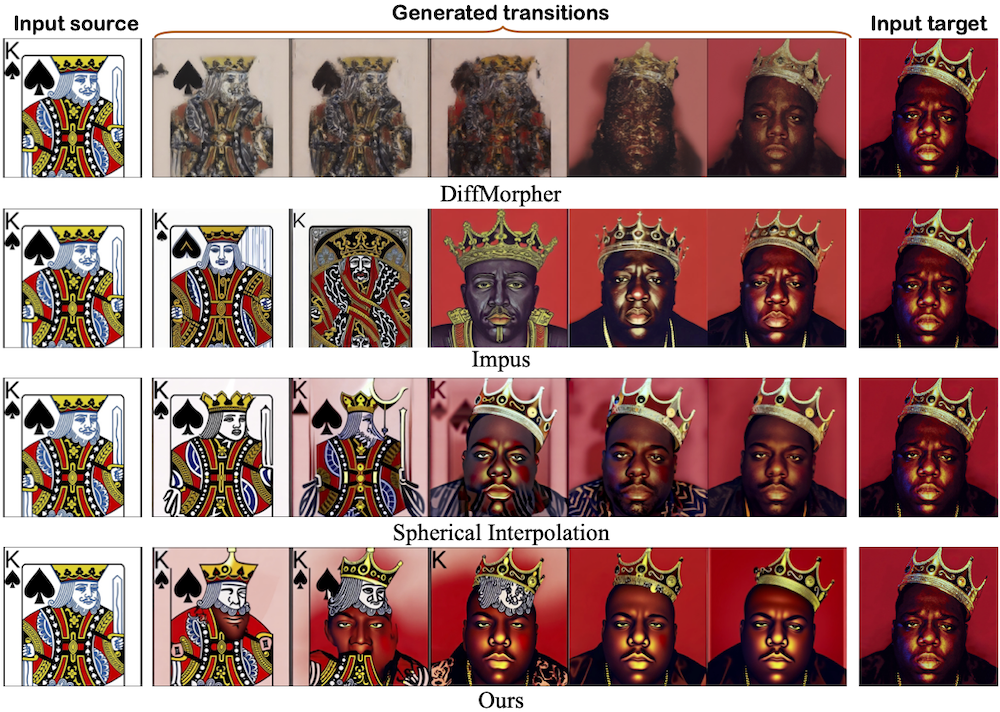}
  \caption[More qualitative comparisons with existing techniques (Part III).]{\textbf{More qualitative comparisons with existing techniques (Part III).} }
   \label{fig:more-comparison3}
\end{figure*} 

\begin{figure*}[h]
  \centering
   \includegraphics[width=0.9\linewidth]{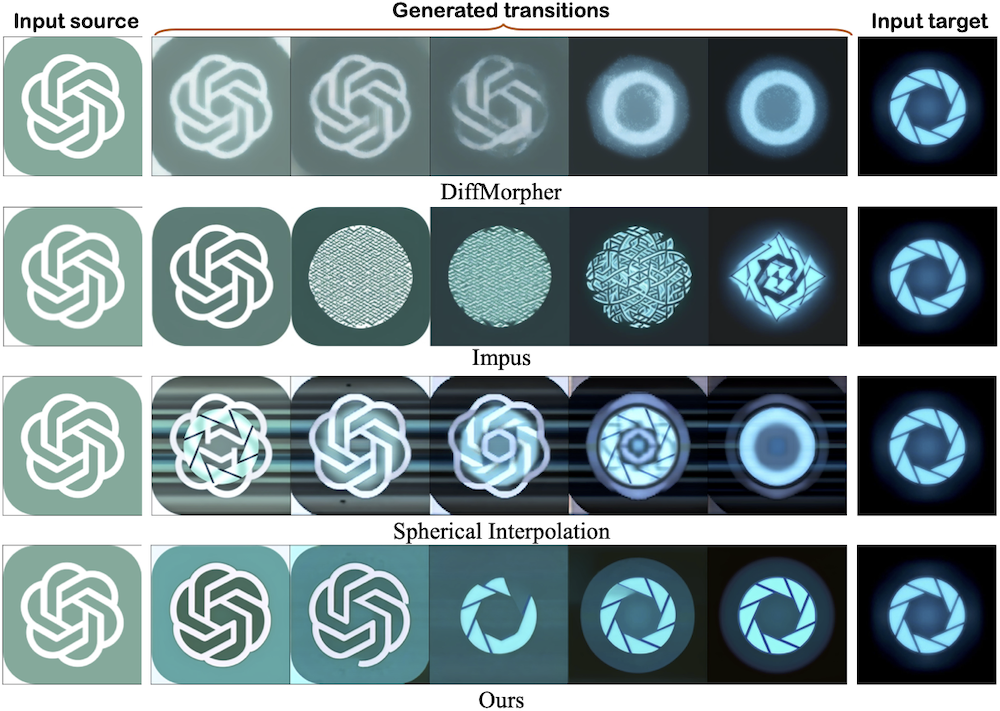}
   \includegraphics[width=0.9\linewidth]{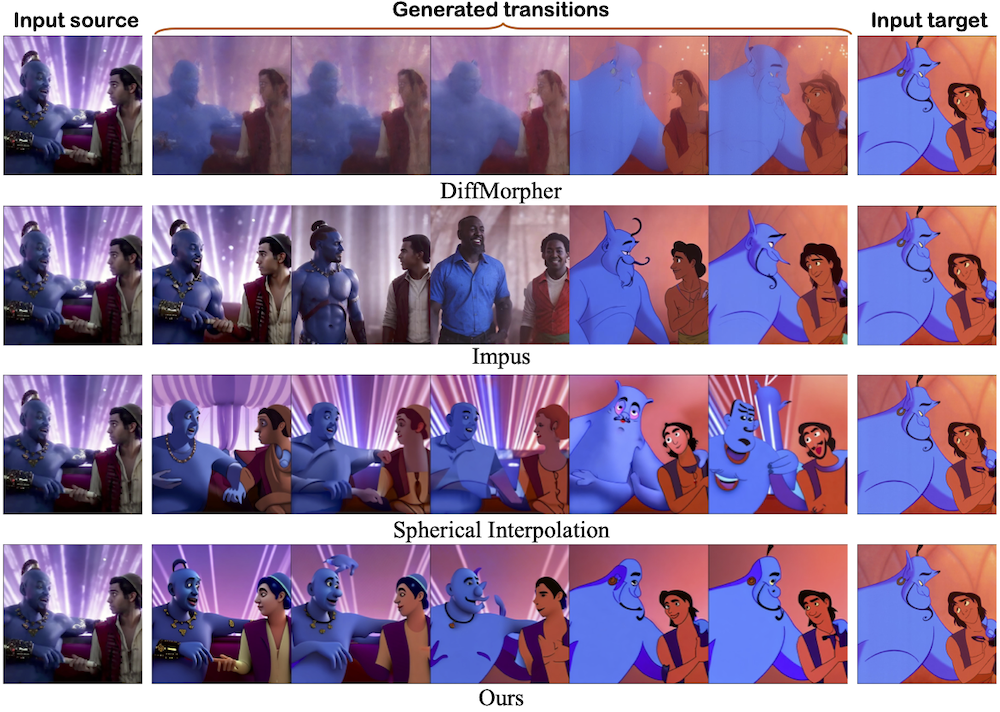}
  \caption[More qualitative comparisons with existing techniques (Part IV).]{\textbf{More qualitative comparisons with existing techniques (Part IV).} }
   \label{fig:more-comparison4}
\end{figure*} 

\begin{figure*}[h]
  \centering
   \includegraphics[width=0.9\linewidth]{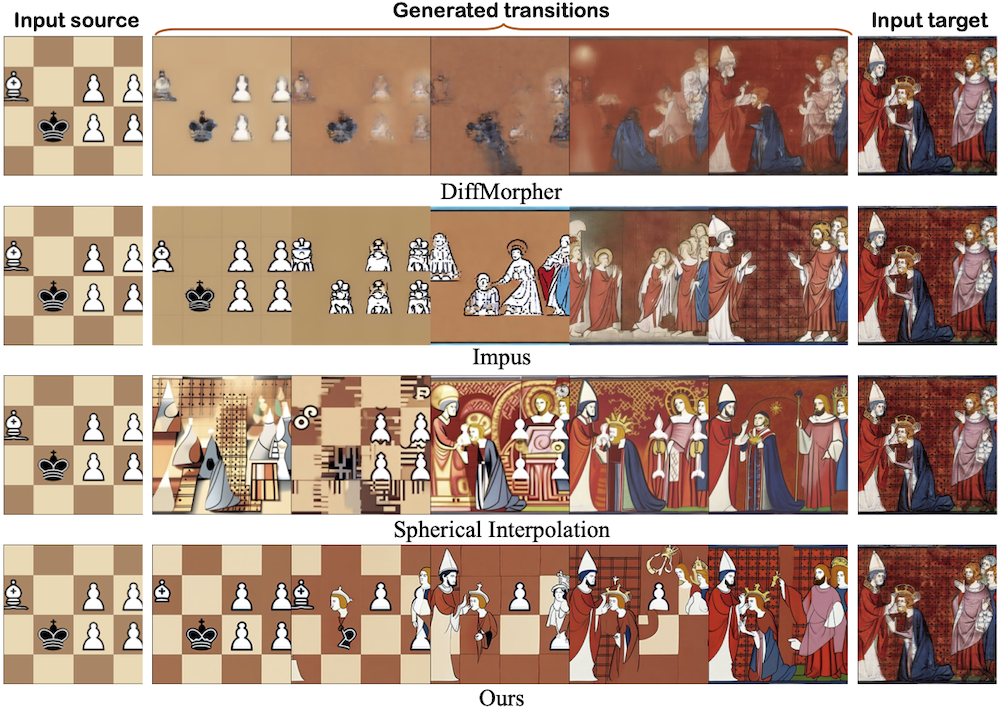}
   \includegraphics[width=0.9\linewidth]{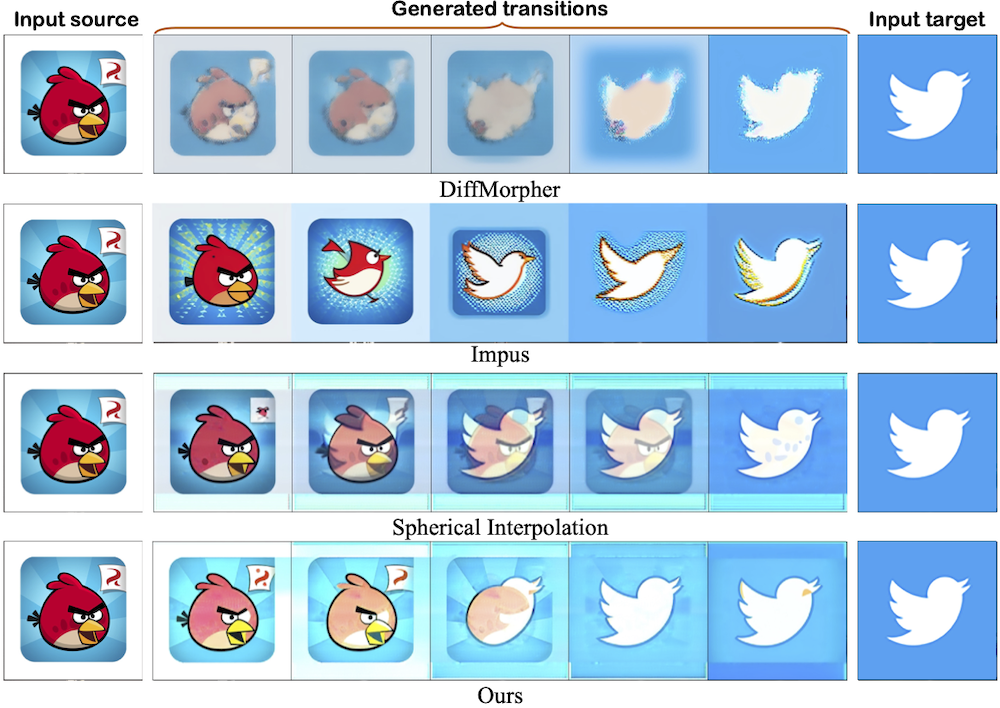}
  \caption[More qualitative comparisons with existing techniques (Part V).]{\textbf{More qualitative comparisons with existing techniques (Part V).} }
   \label{fig:more-comparison5}
\end{figure*} 

\begin{figure*}[h]
  \centering
   \includegraphics[width=0.9\linewidth]{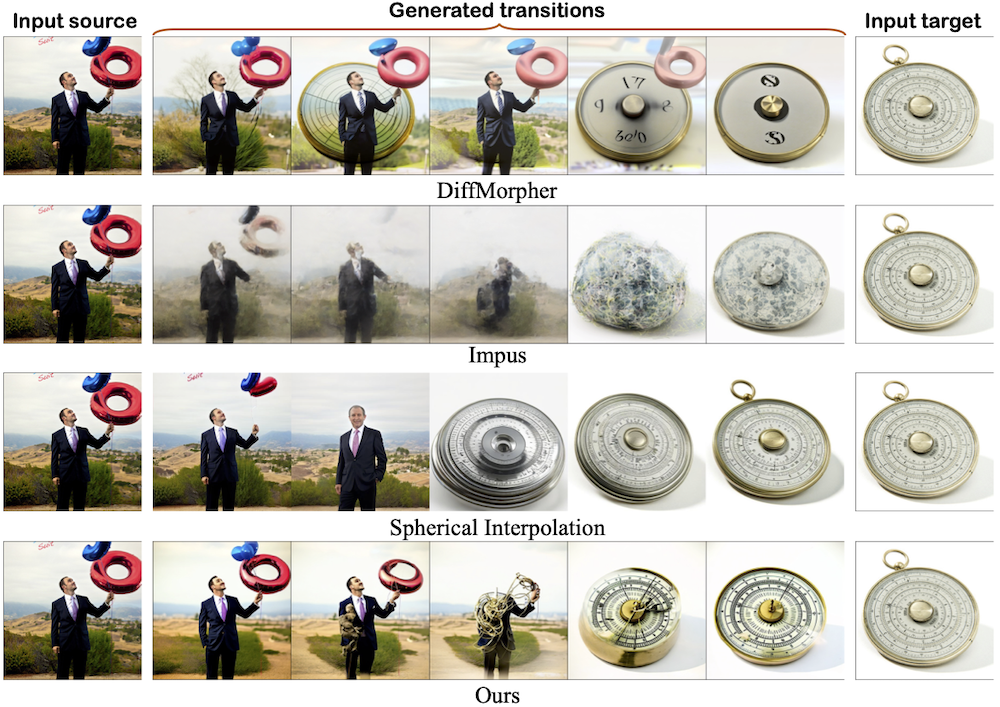}
   \includegraphics[width=0.9\linewidth]{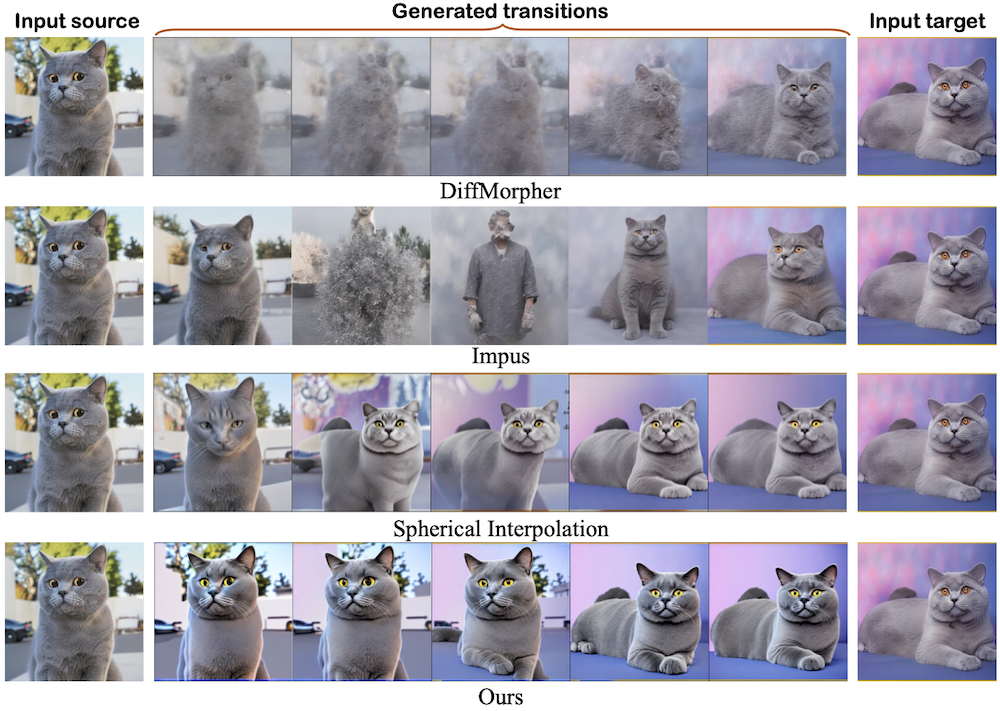}
  \caption[More qualitative comparisons with existing techniques (Part VI).]{\textbf{More qualitative comparisons with existing techniques (Part VI).} }
   \label{fig:more-comparison6}
\end{figure*} 

\begin{figure*}[h]
  \centering
   \includegraphics[width=0.9\linewidth]{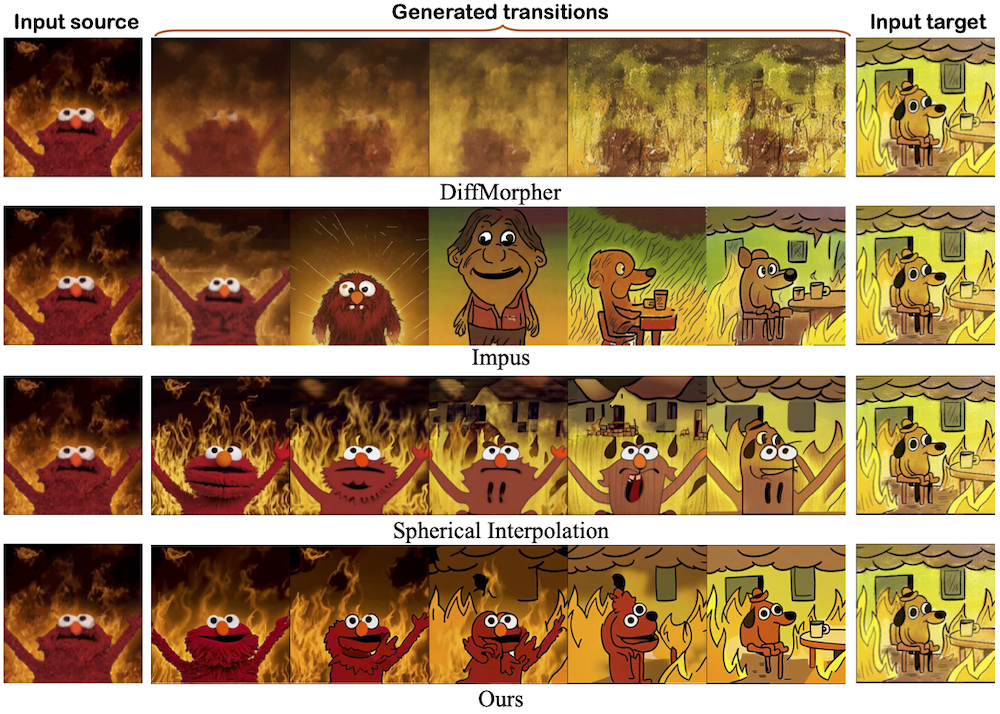}
   \includegraphics[width=0.9\linewidth]{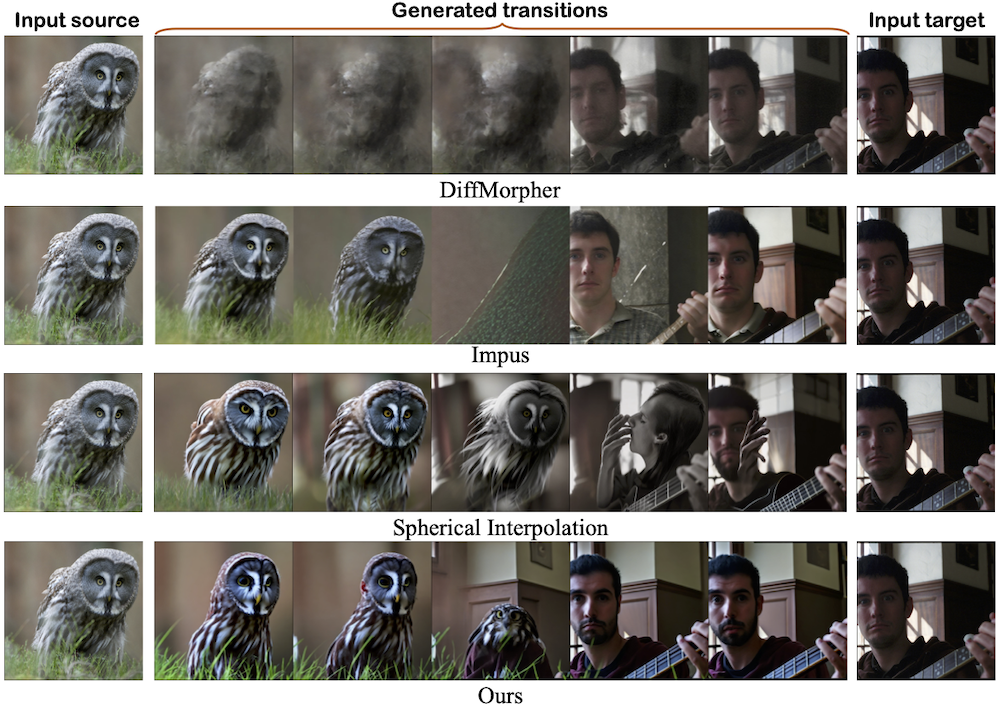}
  \caption[More qualitative comparisons with existing techniques (Part VII).]{\textbf{More qualitative comparisons with existing techniques (Part VII).} }
   \label{fig:more-comparison7}
\end{figure*} 

\begin{figure*}[h]
  \centering
   \includegraphics[width=0.9\linewidth]{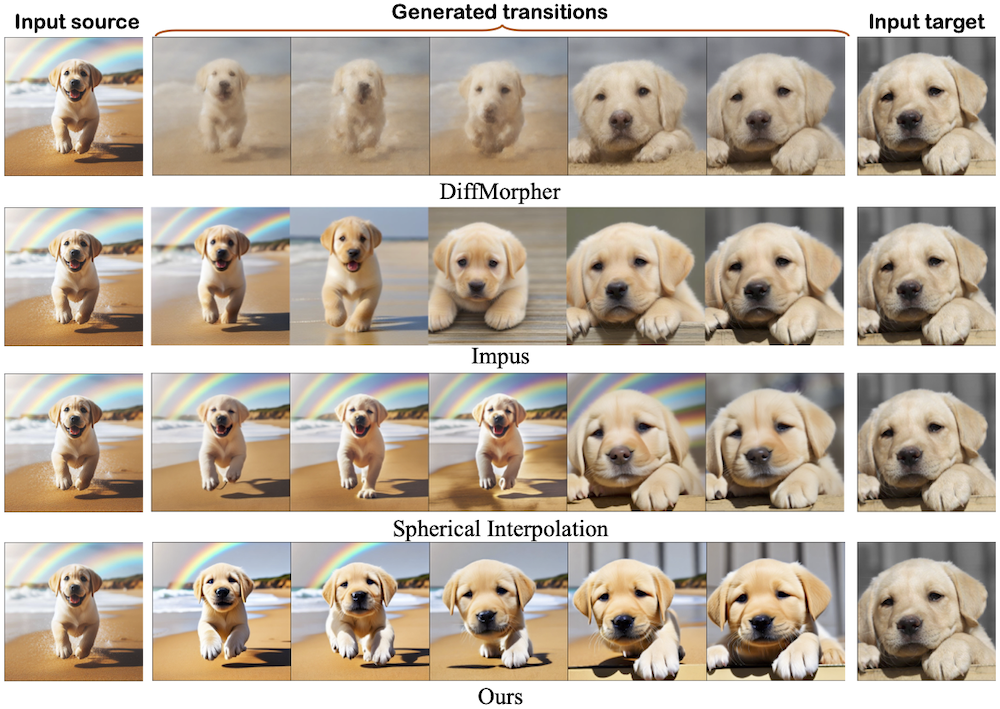}
   \includegraphics[width=0.9\linewidth]{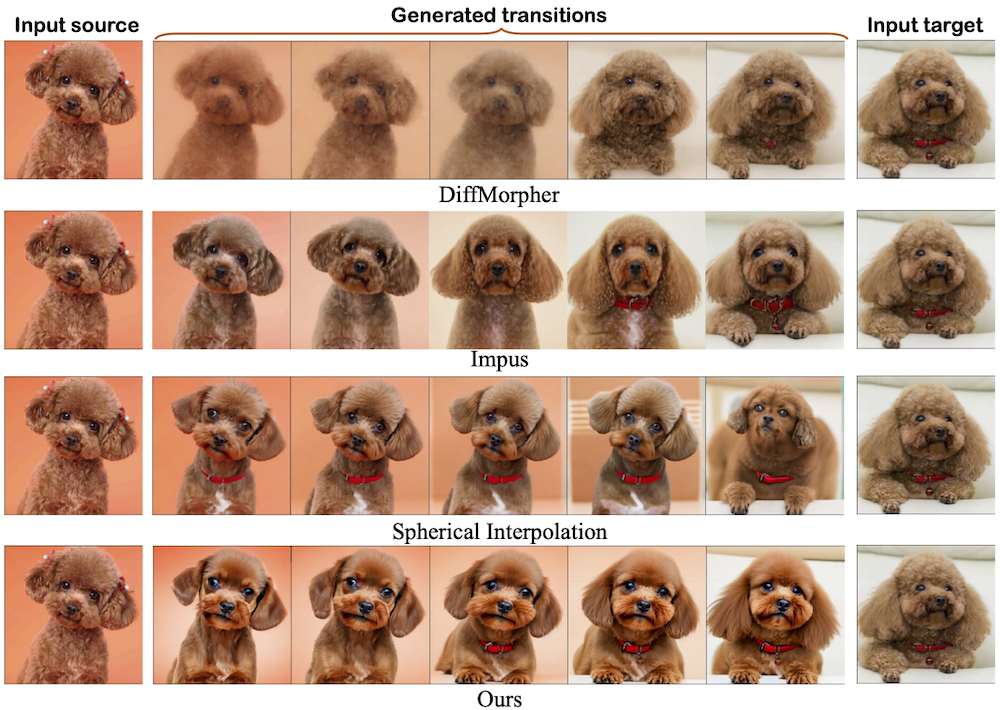}
  \caption[More qualitative comparisons with existing techniques (Part VIII).]{\textbf{More qualitative comparisons with existing techniques (Part VIII).} }
   \label{fig:more-comparison8}
\end{figure*} 

%% file: Figures/supp/qualitative-results.tex
\begin{figure*}[h]
  \centering
   \includegraphics[width=0.9\linewidth]{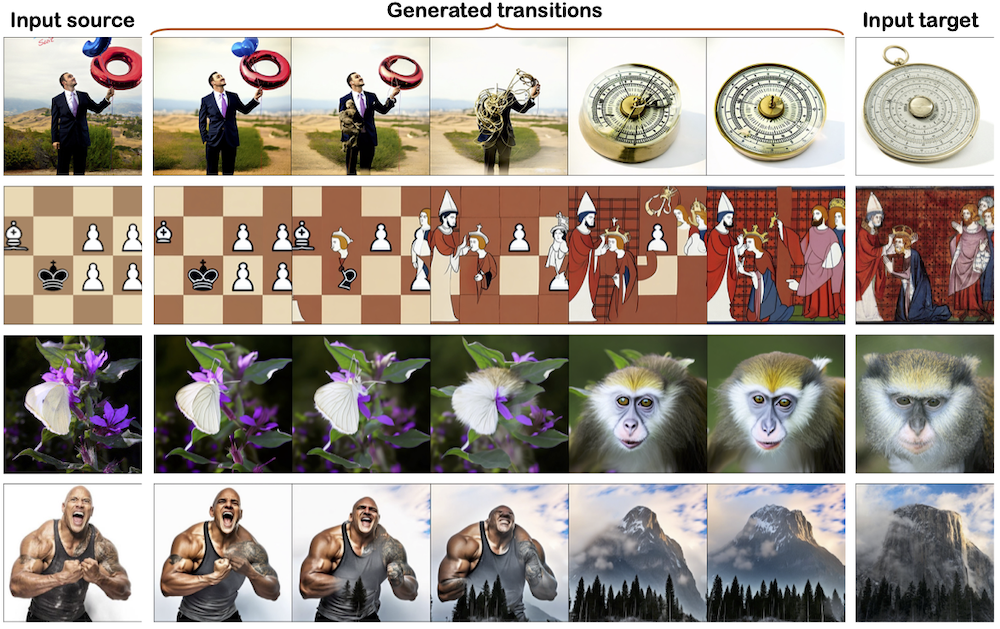}
   \includegraphics[width=0.9\linewidth]{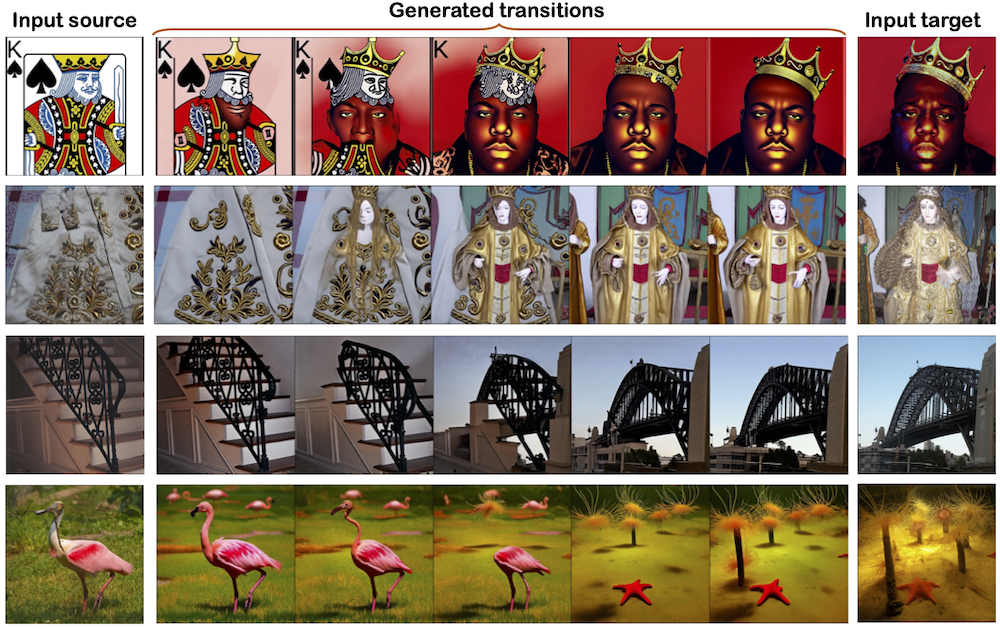}
  \caption[Images with different semantics and different layouts.]{\textbf{Images with different semantics and different layouts.} }
   \label{fig:more-results1}
\end{figure*} 

\begin{figure*}[h]
  \centering
   \includegraphics[width=0.9\linewidth]{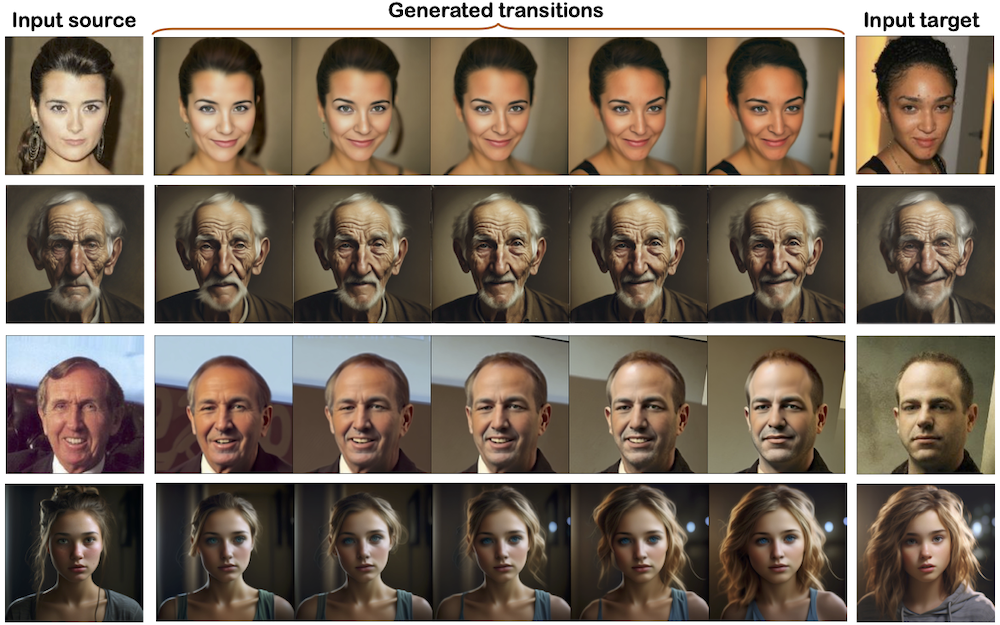}
   \includegraphics[width=0.9\linewidth]{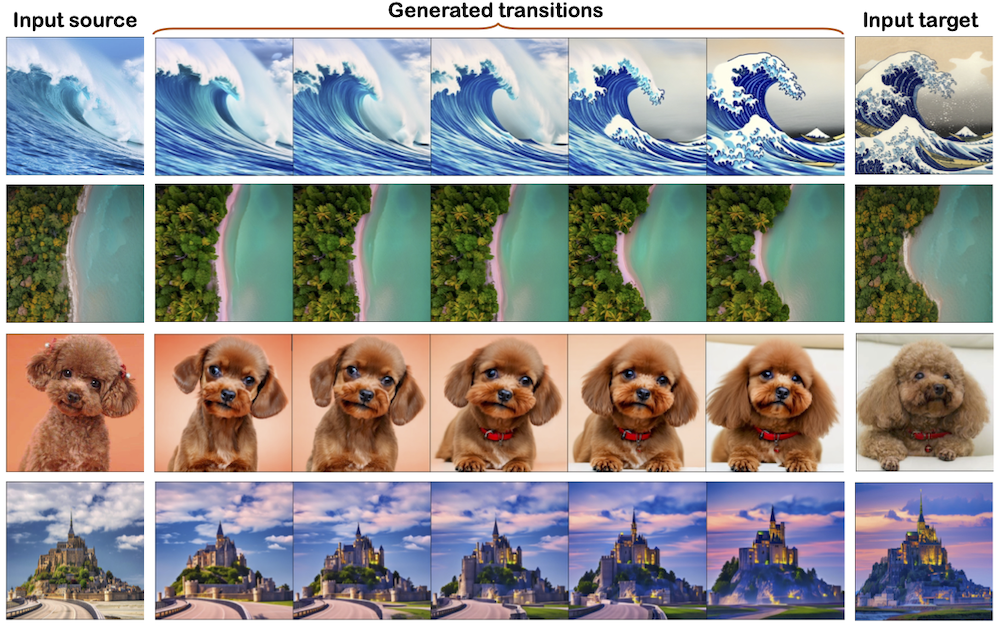}
  \caption[Images with similar semantics and similar layouts.]{\textbf{Images with similar semantics and similar layouts.} }
   \label{fig:more-results2}
\end{figure*} 

\begin{figure*}[h]
  \centering
   \includegraphics[width=0.9\linewidth]{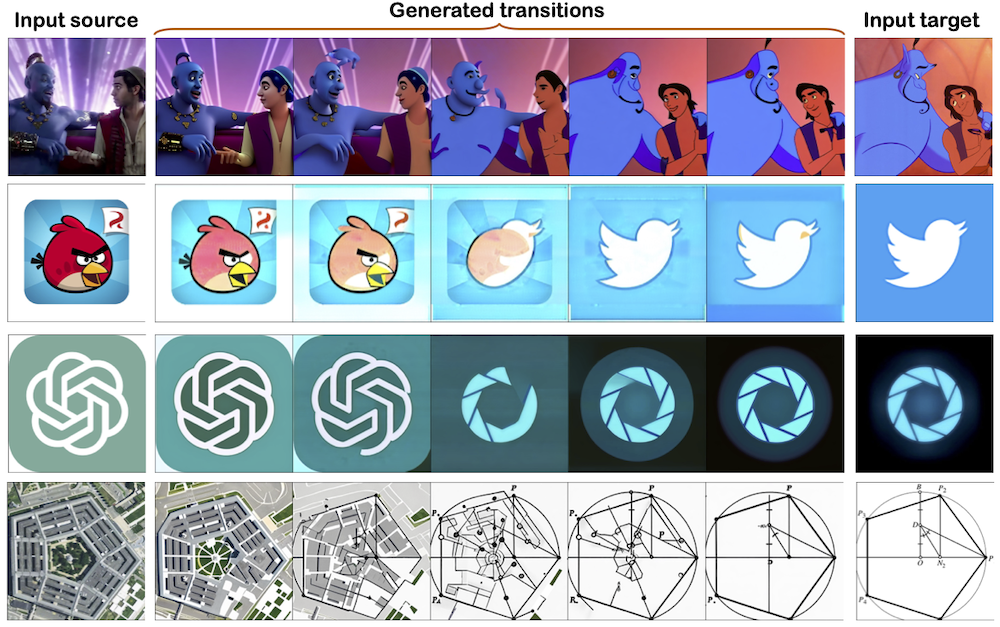}
  \caption[Images with different semantics and similar layouts.]{\textbf{Images with different semantics and similar layouts.} }
   \label{fig:more-results3}
\end{figure*} 

\begin{figure*}[h]
  \centering
   \includegraphics[width=0.9\linewidth]{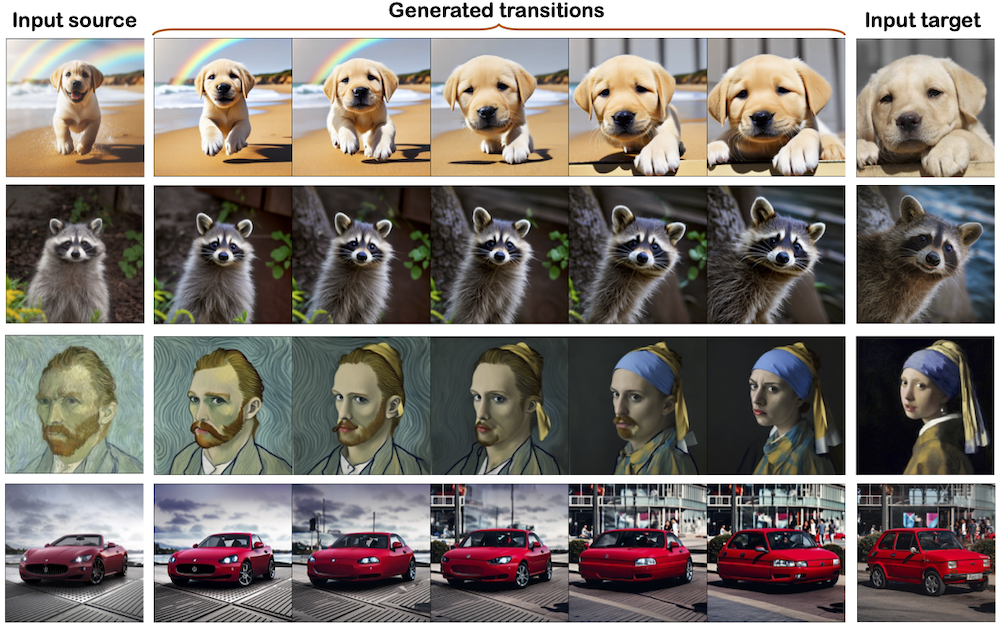}
  \caption[Images with similar semantics and different layouts.]{\textbf{Images with similar semantics and different layouts.} }
   \label{fig:more-results4}
\end{figure*}

%% file: Figures/supp/morph4data.tex
\begin{figure*}[h]
  \centering
   \includegraphics[width=\linewidth]{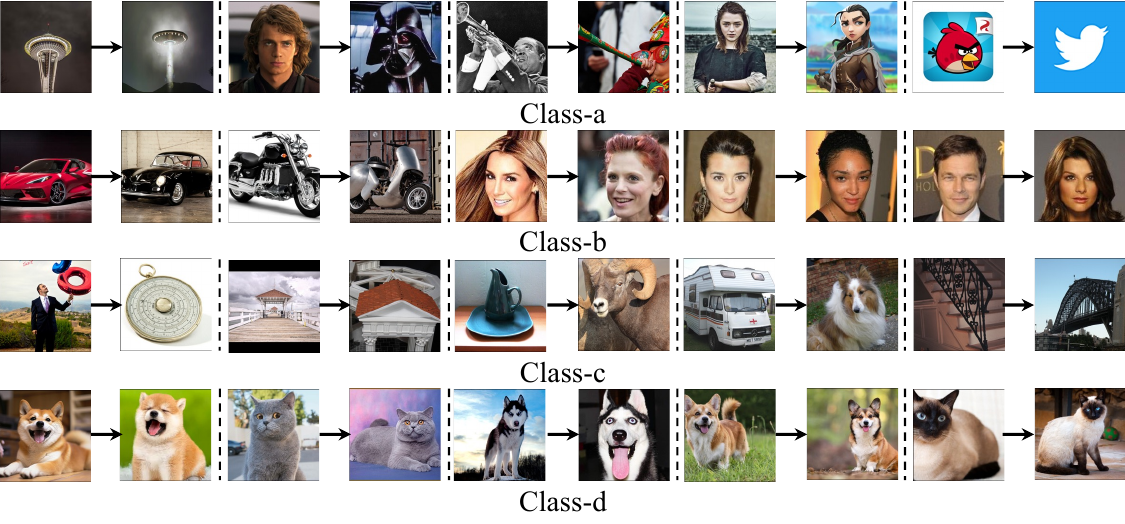}
  \caption[Examples of 4 classes in Morph4Data.]{\textbf{Examples of 4 classes in Morph4Data.} }
   \label{fig:data}
\end{figure*} 

%% file: Figures/supp/application.tex
\begin{figure*}[h]
  \centering
   \includegraphics[width=\linewidth]{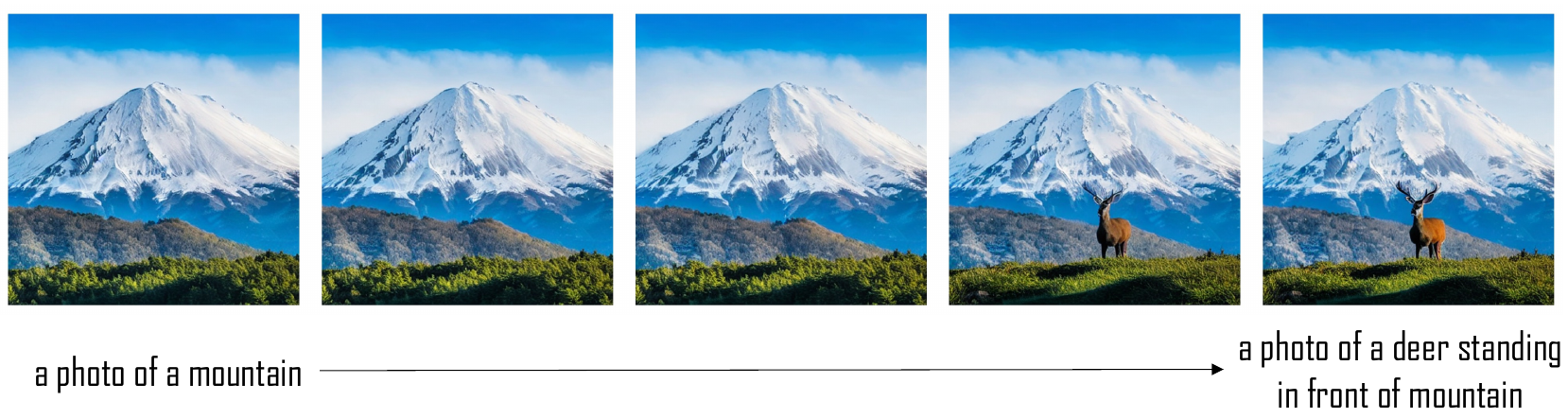}
  \caption[Application of \OMO in image editing]{\textbf{Application of \OMO in image editing} }
   \label{fig:editing}
\end{figure*} 

%% file: Figures/failure.tex
\begin{figure}[h]
  \centering
   \includegraphics[width=0.7\linewidth]{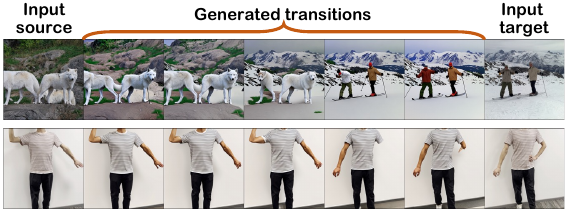}
  \caption[Failure cases]{\textbf{Failure cases.}}
   \label{fig:failure}
\end{figure}